\documentclass{article}

\usepackage{microtype}
\usepackage{amsmath}
\usepackage{amssymb}
\usepackage{graphicx}
\usepackage{subfigure}
\usepackage{booktabs} % for professional tables

\usepackage{hyperref}

\usepackage[accepted]{icml2019}

\icmltitlerunning{Understanding the One-pixel Attack: Propagation Maps and Locality Analysis}

\begin{document}

\twocolumn[
\icmltitle{Understanding the One-pixel Attack: Propagation Maps and Locality Analysis}

% It is OKAY to include author information, even for blind
% submissions: the style file will automatically remove it for you
% unless you've provided the [accepted] option to the icml2019
% package.

% List of affiliations: The first argument should be a (short)
% identifier you will use later to specify author affiliations
% Academic affiliations should list Department, University, City, Region, Country
% Industry affiliations should list Company, City, Region, Country

% You can specify symbols, otherwise they are numbered in order.
% Ideally, you should not use this facility. Affiliations will be numbered
% in order of appearance and this is the preferred way.
\icmlsetsymbol{equal}{*}

\begin{icmlauthorlist}
\icmlauthor{Danilo Vasconcellos Vargas}{ky}
\icmlauthor{Jiawei Su}{sky}
\end{icmlauthorlist}

\icmlaffiliation{ky}{Faculty of Information Science and Electrical Engineering, Kyushu University, Fukuoka, Japan}
\icmlaffiliation{sky}{Graduate School of Information Science and Electrical Engineering, Kyushu University, Fukuoka, Japan}

\icmlcorrespondingauthor{Danilo Vasconcellos Vargas}{vargas@inf.kyushu-u.ac.jp}

% You may provide any keywords that you
% find helpful for describing your paper; these are used to populate
% the "keywords" metadata in the PDF but will not be shown in the document
\icmlkeywords{One-Pixel Attack, Adversarial Machine Learning, Understanding Deep Neural Networks, Visualizing Deep Neural Networks, Computer Vision, Machine Learning}

\vskip 0.3in
]

% this must go after the closing bracket ] following \twocolumn[ ...

% This command actually creates the footnote in the first column
% listing the affiliations and the copyright notice.
% The command takes one argument, which is text to display at the start of the footnote.
% The \icmlEqualContribution command is standard text for equal contribution.
% Remove it (just {}) if you do not need this facility.

\printAffiliationsAndNotice{}  % leave blank if no need to mention equal contribution
%\printAffiliationsAndNotice{\icmlEqualContribution} % otherwise use the standard text.

\begin{abstract}
Deep neural networks were shown to be vulnerable to single pixel modifications.
However, the reason behind such phenomena has never been elucidated.
Here, we propose Propagation Maps which show the influence of the perturbation in each layer of the network.
Propagation Maps reveal that even in extremely deep networks such as Resnet, modification in one pixel easily propagates until the last layer.
In fact, this initial local perturbation is also shown to spread becoming a global one and reaching absolute difference values that are close to the maximum value of the original feature maps in a given layer.
%It may be used as a general security check for neural networks.
%Moreover, we do a locality analysis in which we show that (a) nearby pixels of the perturbed one in the one-pixel attack tend to share the same vulnerability and (b) attacked classes show a probabilistic pattern. Both (a) and (b) can be exploited to create a more dangerous near one-shot one-pixel attack.
Moreover, we do a locality analysis in which we demonstrate that nearby pixels of the perturbed one in the one-pixel attack tend to share the same vulnerability, revealing that the main vulnerability lies in neither neurons nor pixels but receptive fields.
%and (b) attacked classes show a probabilistic pattern. Both (a) and (b) can be exploited to create a more dangerous near one-shot one-pixel attack.
Hopefully, the analysis conducted in this work together with a new technique called propagation maps shall shed light into the inner workings of other adversarial samples and be the basis of new defense systems to come. 
\end{abstract}

\section{Introduction}
\label{intro}

\begin{figure}[htb]
\vskip 0.2in
\begin{center}
%centerline{
\includegraphics[width=\columnwidth]{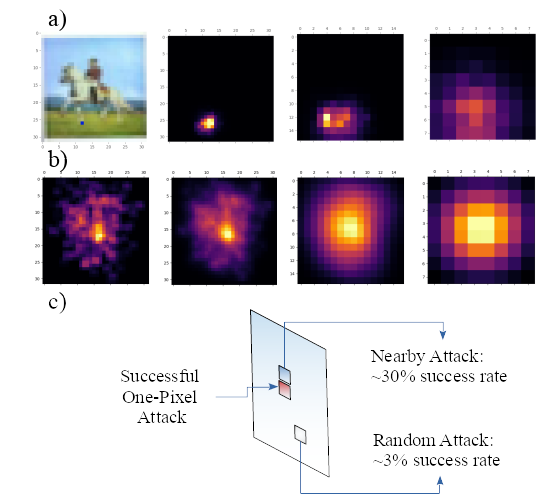}
\caption{\textbf{a)} Propagation Maps (PMmax) of a successful one-pixel attack on Resnet shows how the influence of one pixel perturbation grows and spreads (bright colors show differences in feature map that are close to the maximum original layer output).
\textbf{b)} Average Propagation Map over the entire set of propagation maps shows the overall distribution of attacks and their propagation.
\textbf{c)} Illustration of locality analysis.
}
\end{center}
\vskip -0.2in
\label{overviewpic}
\end{figure}

Recently, a series of papers have shown that deep neural networks (DNN) are vulnerable to various types of attacks \cite{3},\cite{7},\cite{moosavi2017universal},\cite{brown2017adversarial},\cite{su2017one}, \cite{moosavi2016deepfool}.\cite{carlini2017towards},\cite{kurakin2016adversarial},\cite{sharif2016accessorize}, \cite{athalye2017synthesizing} 
However, the reasons underlying these vulnerabilities are still largely unknown.
We argue that one of the most important facets of adversarial machine learning resides in its investigative nature.
In other words, adversarial machine learning provides us with key tools to understand current DNNs.
As much as attacks tells us about the security behind DNNs, they also tells us to what extent DNNs can reason over data and what do they understand to be, for example, the concept of a "car" or "horse".

In this paper, inspired by recent attacks and defenses we propose a technique called propagation maps that would be able to explain most of them.
Here, giving the constrained space, we focus on one attack which is puzzling and largely unexplained, the one-pixel attack.
Propagation maps enable us to show how one pixel perturbation may grow in influence over the layers and spread over many pixels to cause a final change in class.
Moreover, statistical properties of the propagation reveal many properties of the attacks as well as their distribution (Figure~\ref{overviewpic}).

Additionally, to further understand the one-pixel attack a locality analysis is performed.
The locality analysis consists of executing the attack in nearby pixels of a successful one-pixel attack, i.e., using the same pixel perturbation but different pixel position. 
Indeed, the success rate of nearby pixel is effective and equal among different neural networks (Figure~\ref{overviewpic}), showing that rather than pixels or neurons, the vulnerability lies in some of the receptive fields.
This reveals an interesting property shared among DNNs which is independent of the model or attack success rate.
%Consequently, perhaps a better name to one-pixel attack is receptive field attack.

\section{Related Work}

In the following subsections an overview of different types of attacks and a brief overview of the related works will be covered.

\subsection{Adversarial Samples and Different Types of Attacks}

The samples that can make machine learning algorithms misclassify received the name of adversarial samples.
Let $f(\textbf{x}) \in \mathbb{R}^k$ be the output of a machine learning algorithm in which $\textbf{x} \in \mathbb{R}^{m \times n}$ is the input of the algorithm for input and output of sizes $m \times n$ and $k$ respectively.
It is possible to define adversarial samples $\textbf{x'}$ explicitly as follows:
\begin{equation}
\begin{aligned}
 & \textbf{x'} = \textbf{x} + \epsilon_{\textbf{x}} \\
 & \{ \textbf{x'} \in \mathbb{R}^n \mid \underset{j}{\operatorname{argmax}}(f(\textbf{x'})_j) \ne \underset{i}{\operatorname{argmax}}(f(\textbf{x})_i),
\end{aligned}
\end{equation}
in which $\epsilon_{\textbf{x}} \in \mathbb{R}^{m \times n}$ is a small perturbation added to the input.

In adversarial machine learning one want to search for adversarial samples. 
For that it is possible to use the knowledge of the DNN in question to craft samples such as using back-propagation for obtaining gradient information and subsequently using gradient descent as done by the ``fast gradient sign'' proposed by I.J. Goodfellow et al. \cite{2}.
There is also the greedy perturbation searching method proposed by S.M. Moosavi-Dezfooli et al. \cite{moosavi2016deepfool} and N. Papernot et al. utilize Jacobian matrix of the function learned during training to create a saliency map which will guide the search for adversarial samples \cite{1}. 

However, it is possible to search for adversarial samples without taking into account the internal characteristics of DNNs.
This type of model agnostic search is also called black-box attack.
To search for adversarial samples in this scenario, it is common to use perturbations that increase a given soft label $f(\textbf{x})_t$ (targetted attack) in which $t$ is the index of the target class. 
In other words, they can be defined as the following optimization problem:
\begin{equation}
\begin{aligned}
& \underset{\epsilon_{\textbf{x}}}{\text{maximize}}
& & f(\textbf{x}+\epsilon_{\textbf{x}})_t \\
& \text{subject to}
& & \Vert \epsilon_{\textbf{x}} \Vert \leq L 
%& & \Vert e(\textbf{x}) \Vert_{0} \leq d
\end{aligned}
\end{equation}

Regarding untargeted attacks, the objective function can be defined, for example, as the minimization of the soft label for the outputted class $f(\textbf{x})_i$. 
See the complete equation below. 
\begin{equation}
\begin{aligned}
& \underset{\epsilon_{\textbf{x}}}{\text{maximize}}
& & -f(\textbf{x}+\epsilon_{\textbf{x}})_i \\
& \text{subject to}
& & \Vert \epsilon_{\textbf{x}} \Vert \leq L 
%& & \Vert e(\textbf{x}) \Vert_{0} \leq d
\end{aligned}
\label{untargeted_attack}
\end{equation}
The minimization of the difference between the highest soft-label index and the second highest one is also one of the other possibilities of objective function for untargeted attacks. 
There are many black-box attacks in the literature. 
To cite some \cite{13},\cite{24},\cite{25}. 

%There have been many efforts to understand DNN by visualizing the activation of network nodes \cite{16},\cite{18},\cite{19},\cite{20} while the geometrical characteristics of DNN boundary have gained less attraction due to the difficulty of understanding high-dimensional space. 
%However, the robustness evaluation of DNN with respect to adversarial perturbation might shed light in this complex problem \cite{27}. 
%For example, both natural and random images are found to be vulnerable to adversarial perturbation. 
%Assuming these images are evenly distributed, it suggests that most data points in the input space are gathered near to the boundaries \cite{27}. 
%In addition, A. Fawzi et al. revealed more clues by conducting a curvature analysis. 
%Their conclusion is that including the connectivity of regions of the same class, 
%Their conclusion is that the region along most directions around natural images are flat with only few directions where the space is curved and the images are sensitive to perturbation \cite{26}.  
%Interestingly, universal perturbations (i.e. a perturbation that when added to any natural image can generate adversarial samples with high effectiveness) were shown possible and to achieve a high effectiveness when compared to random pertubation. 
%This indicates that the diversity of boundaries might be low while the boundaries' shapes near different data points are similar \cite{28}. 

It is important to note that $\epsilon$ should be small enough to not allow an image to become a different class.
Such a transformation would invalidate the adversarial sample creation because it is not a misclassification.
Since most attacks use perturbations which comprise of the whole image, $\Vert \epsilon_{\textbf{x}} \Vert \leq L $ is a good optimization constraint.
However, it is also possible to look the other way around and deal with few perturbed dimensions.
In this case, the constraint changes to a L0 norm which actually counts the dimensions of the the perturbation, i.e., the total number of non-zero elements in the perturbation vector.
The complete equation is as follows:

\begin{equation}
\begin{aligned}
& \underset{e(\textbf{x})^{*}}{\text{maximize}}
& & f_{adv}(\textbf{x}+e(\textbf{x})) \\
& \text{subject to}
%& & \Vert e(\textbf{x}) \Vert \leq L
& & \Vert e(\textbf{x}) \Vert_{0} \leq d,
\end{aligned}
\end{equation}
where $d$ is a small number of dimensions ($d=1$ for the one-pixel attack).

\subsection{Recent Advances in Attacks and Defenses}

The question of if machine learning is secure was asked some time ago \cite{6},\cite{barreno2010security}.
However, it was only in 2013 that Deep Neural Networks' (DNN) security was completely put into question \cite{3}.
C. Szegedy et al. demonstrated that by adding noise to an image it is possible to produce a visually identical image which can make DNNs misclassify.
This was conterintuitive, since the DNNs that misclassified had a very high accuracy in the tests rivaling even the accuracy of human beings.
%However, they were not capable of correctly recognizing images with small noise.
%In other words, they had learned some model that although accurate did not use 

Recently, the vulnerabilities of neural networks were shown to be even more aggravating.
For example, A. Nguyen et al. did a series of experiments in which images were continuously developed to fool a machine learning algorithm. 
They found out that DNNs give high confidence results to random noise.
DNNs also tend to classify patterns or textures, such as the stripes of a bus as a bus, with high confidency \cite{7}. 
Universal adversarial perturbations in which a single crafted perturbation is able to make a DNN misclassify multiple samples was also shown to be possible \cite{moosavi2017universal}. 
In \cite{brown2017adversarial}, the authors showed that it is possible to make DNNs misclassify by adding a patch to an image.
Interestingly, each patch is a targetted attack, i.e., a banana patch would make a DNN misclassify the image as a banana when added.
Moreover, in \cite{su2017one} it was shown that even one pixel could make DNNs' misclassify. 
Indicating that although DNNs have a high accuracy in recognition tasks, their "understanding" of what is a "dog" or "cat" is still very different from human beings.
In fact, adversarial samples can be used to evaluate the robustness of a DNN \cite{moosavi2016deepfool}.\cite{carlini2017towards}.

%Real world scenarios with varying light conditions among other elements present new difficulties for attacks to succed.
Although much of the research in adversarial machine learning is conducted under ideal conditions in a laboratory, the same techniques are not difficult to apply to real world scenarios because printed out adversarial samples still work, i.e., many adversarial samples are robust against different light conditions  \cite{kurakin2016adversarial}.
In \cite{sharif2016accessorize}, it was show that it is possible to build glasses that could fool DNNs to believe a given person is somebody else. 
The glasses were custom made and had some orientation in mind when designed.
In fact, in \cite{athalye2017synthesizing} the authors go a step further and verify the existence of 3d adversarial objects which can fool DNNs even when viewpoint, noise, and different light conditions are taking into consideration.

Defensive systems and detection systems were soon proposed to mitigate some of the problems.
Regarding defensive systems, defensive distillation in which a smaller neural network squeezes the content learned by the original one was proposed as a defense \cite{papernot2016distillation} however it was soon shown to not be robust enough \cite{carlini2017towards}.
Adversarial training was also proposed in wchi adversarial images are added to the training dataset in such a way that the DNN will be able to correctly classify them, increasing its robustness \cite{goodfellow2014explaining},\cite{huang2015learning}, \cite{madry2017towards}. However this technique is still vulnerable to black-box attacks \cite{tramer2017ensemble}.
There are many recent variations proposed recently
\cite{ma2018characterizing}, 
\cite{buckman2018thermometer},
\cite{guo2017countering},
\cite{dhillon2018stochastic},
\cite{xie2017mitigating},
\cite{song2017pixeldefend},
\cite{samangouei2018defense},
it is out of the scope of this article to review all of them.
Most of these defenses create some sort of gradient masking, called obfuscated gradients. 
They are carefully analyzed and many of their shortcommings are explained in \cite{athalye2018obfuscated} 
%However, the benefits of using these defenses remains unconclusive.

Regarding detection systems, if it is hard to defend perhaps it would be possible to detect attacks.
A study from \cite{grosse2017statistical} revealed that indeed some adversarial samples have different statistical properties.
Moreover, such a detection system would increase the cost of attacks.
In \cite{xu2017feature}, the authors propose an interesting method in which the classifier compares its prediction with a prediction of the same input but "squeezed" (either color or spatial smoothing).
This allow classifiers to detect with high accuracy adversarial samples.
Having said that, many detection systems are subject to fail when adversarial samples differ from test conditions \cite{carlini2017adversarial},\cite{carlini2017magnet}.
Thus, the increase in cost in part of the attacker seems certain but the clear benefits of detection systems remains unconclusive.

There are many works in attacks and defenses but the reason behind such lack of robustness for accurate classifiers is still largely unknown.
In \cite{goodfellow2014explaining} it is argued that DNNs' linearity are one of the main reasons.
%However, this hypothesis was not verified and remains unconclusive.
%However, this fail to explain the one-pixel attack in which a small perturbation seem to have big effects. 
%%%%%%%%%%%%%%%%%%%%%55%Include after%%%%%%%%%%%%%%%%%%%%%%%%%
If this is the case, perhaps hybrid systems that can leverage the non-linearity that arise from complex models by using evolutionary based optimization techniques such as self-organizing classifiers \cite{vargas2013self} and neuroevolution with unified neuron models \cite{vargas2017spectrum} would make for a promising investigation.
%It is also mentioned in many papers that the distance to the decision boundary of adversarial samples is a key point.
%However, in high-dimensional decision spaces a solution or the proof of its existence remains an open field.
%%%%%%%%%%%%%%%%%%%%%%%%%%%%%%%%%%%%%%%%%%%%%%%

%Interestingly, the one-pixel attack reveal an extreme sensitivity to context and very different understanding of concepts like "car" or "dog" from that of human beings.
%Revealing that even suprahuman neural networks fail to grasp the concepts they predict well.
%This raises questions of broader scope about what is it to learn.

\section{One-Pixel Attack}

One-Pixel Attack investigated the opposite extreme of most attacks to date.
Instead of searching for small spreaded perturbation, it focus on perturbing just one pixel.
This vulnerability to one-pixel is a totally different scenario, i.e., neural networks that are vulnerable to usual attacks may not be vulnerable to one-pixel attack and vice-versa.
%In fact, this makes for a different attack vector which neural networks should also be prepared for.

To achieve such an attack in a black-box scenario the authors used differential evolution which is a simple yet effective evolutionary (DE) algorithm \cite{storn1997differential}.
A candidate solution is coded as a pixel position and its related perturbation. 
The DE search for promising candidate solutions by minimizing the output label of the correct class (Equation~\ref{untargeted_attack}).
In this paper, we use the same differential evolution settings as the original paper \cite{su2017one}.
However, here we define a successful attack as an adversarial attack made over a correctly classified sample.
As a consequence, adversarial attacks over already missclassified samples will be ignored.

\section{Propagation Maps}
\begin{figure*}[htb]
\vskip 0.2in
\begin{center}
%centerline{
\includegraphics[width=0.4\columnwidth]{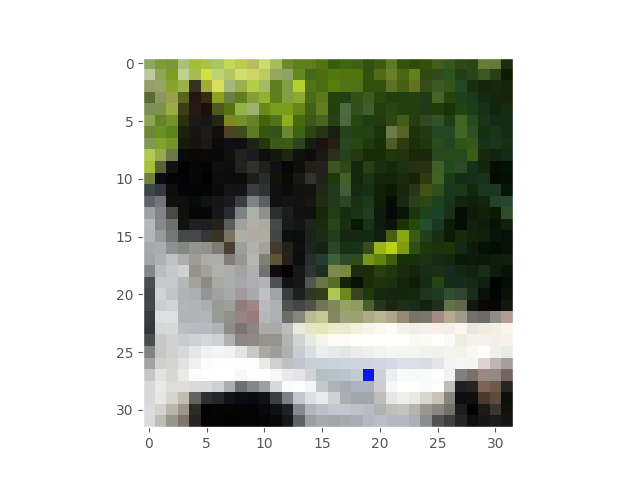}
\includegraphics[width=0.3\columnwidth]{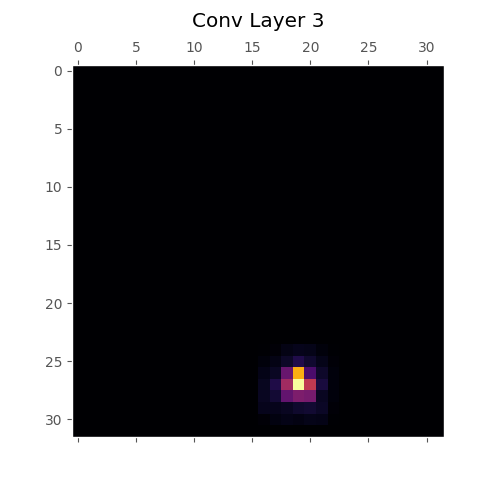}
\includegraphics[width=0.3\columnwidth]{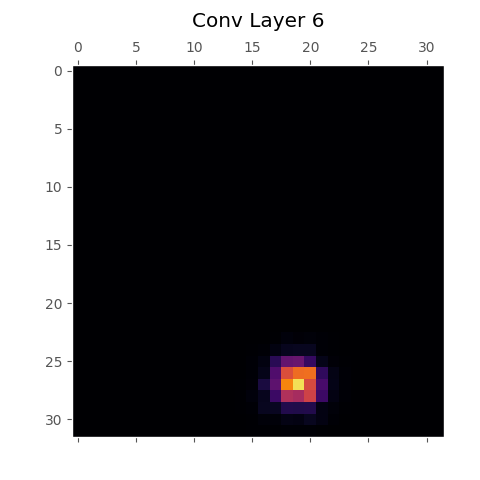}
\includegraphics[width=0.3\columnwidth]{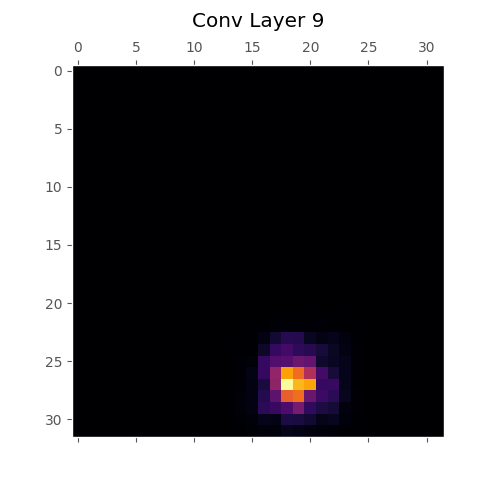}
\includegraphics[width=0.3\columnwidth]{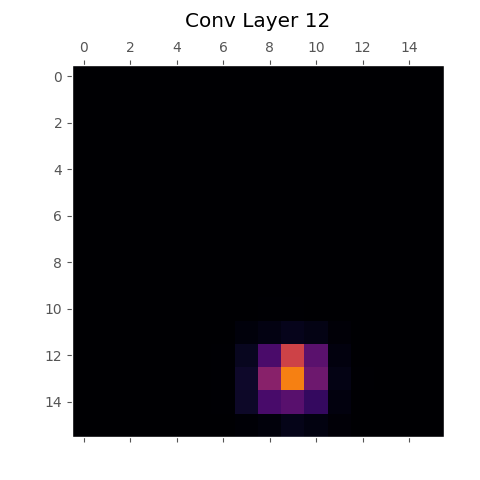}
\includegraphics[width=0.3\columnwidth]{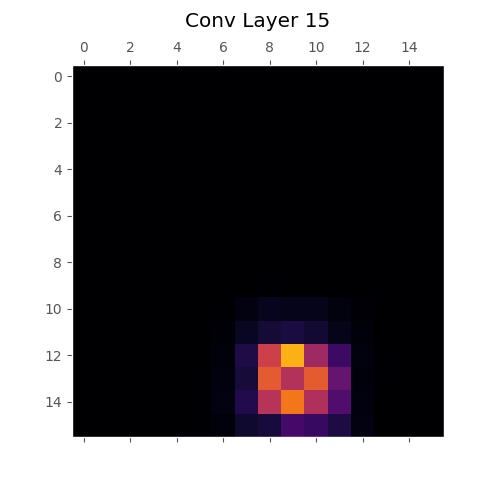}
\includegraphics[width=0.3\columnwidth]{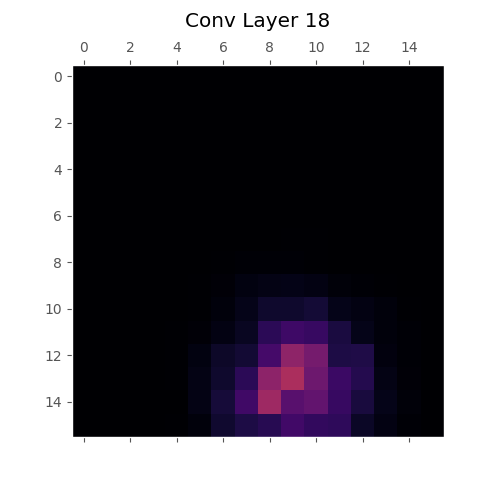}
\includegraphics[width=0.3\columnwidth]{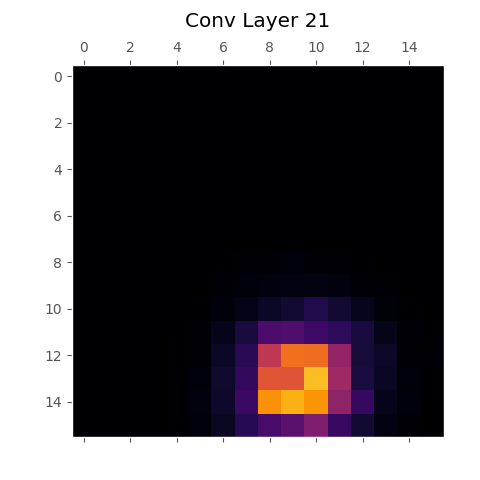}
\includegraphics[width=0.3\columnwidth]{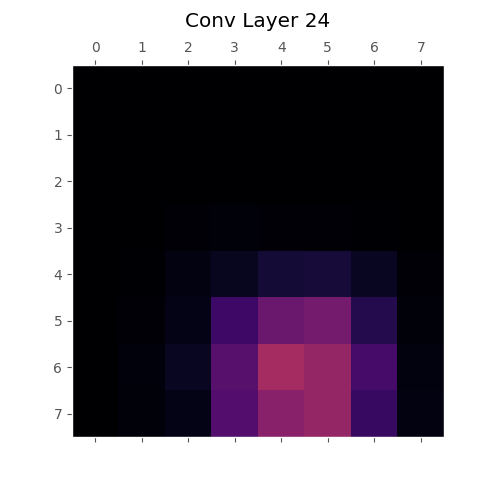}
\includegraphics[width=0.3\columnwidth]{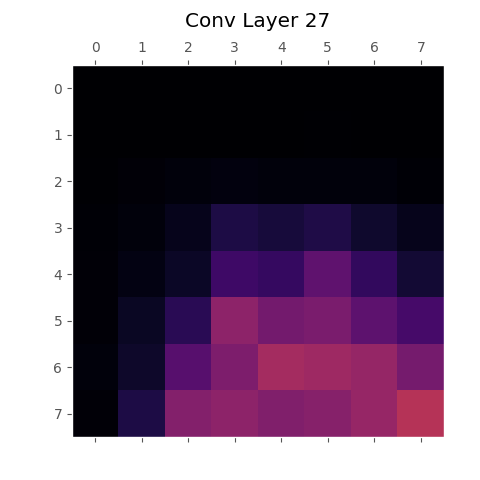}
\includegraphics[width=0.3\columnwidth]{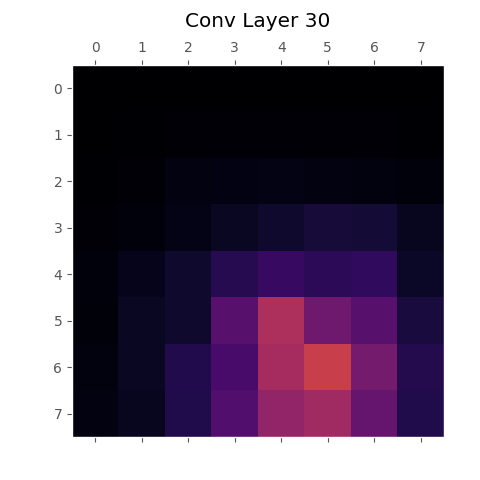}
\includegraphics[width=0.3\columnwidth]{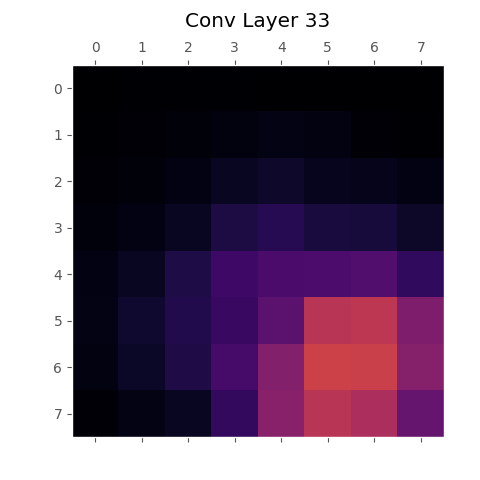}
\caption{Propagation Map (PMmax) for Resnet using a sample from CIFAR.
For this experiment, Equation~\ref{emax} is used. 
The sample above is incorrectly classified as automobile after one pixel is changed in the image.
Values are scaled with the maximum value for each layer of the feature maps being the maximum value achievable in the color map.
Therefore, bright values show that the difference in activation is close to the maximum value in the feature map, demonstrating the strength of the former.
}
\label{advmax}
\end{center}
\vskip -0.2in
\end{figure*}

\begin{figure*}[ht]
\vskip 0.2in
\begin{center}
%centerline{
\includegraphics[width=0.4\columnwidth]{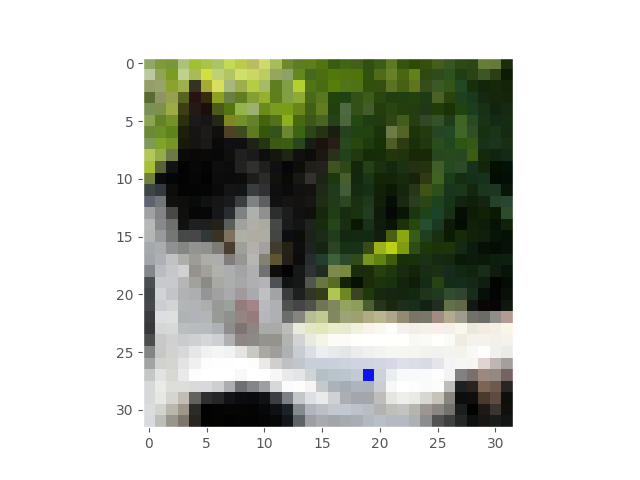}
\includegraphics[width=0.3\columnwidth]{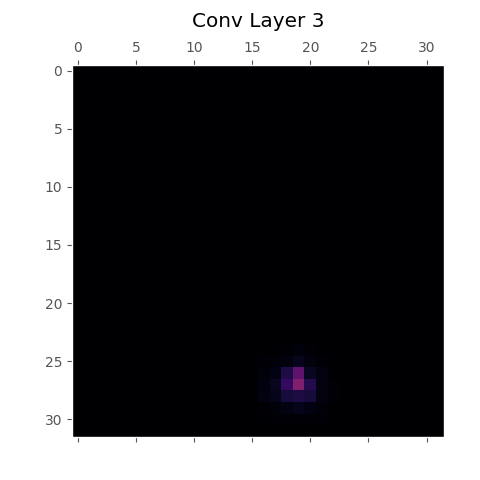}
\includegraphics[width=0.3\columnwidth]{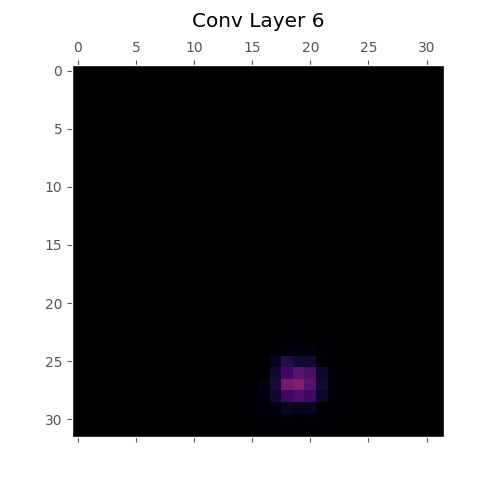}
\includegraphics[width=0.3\columnwidth]{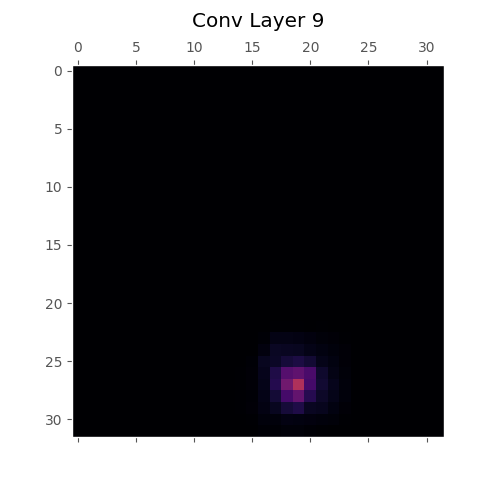}
\includegraphics[width=0.3\columnwidth]{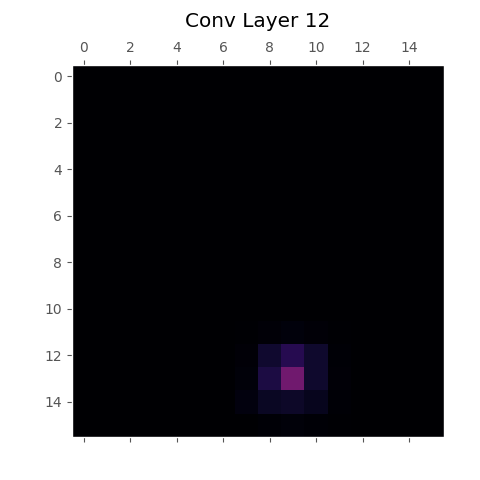}
\includegraphics[width=0.3\columnwidth]{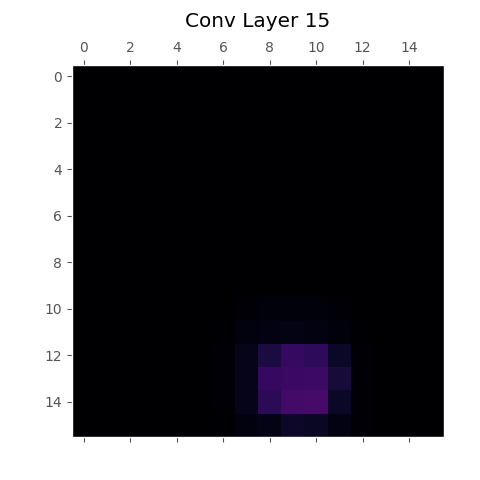}
\includegraphics[width=0.3\columnwidth]{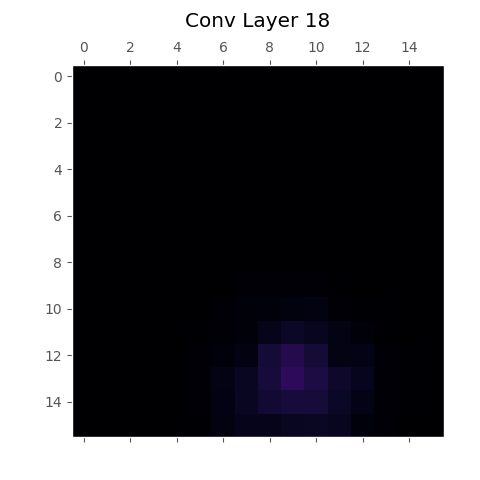}
\includegraphics[width=0.3\columnwidth]{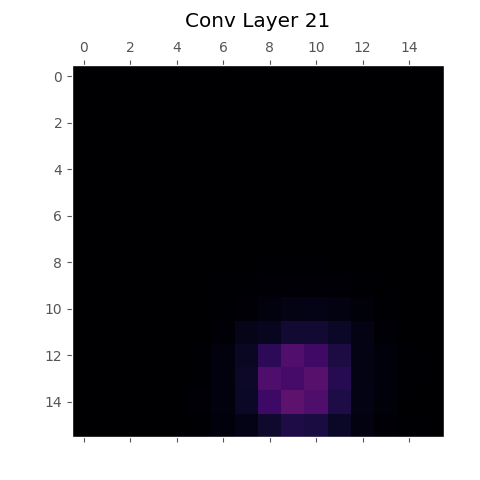}
\includegraphics[width=0.3\columnwidth]{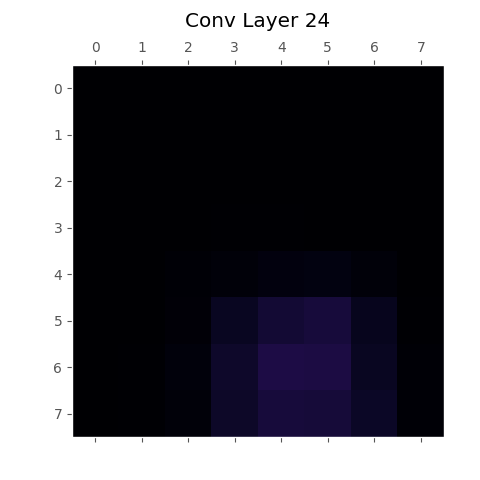}
\includegraphics[width=0.3\columnwidth]{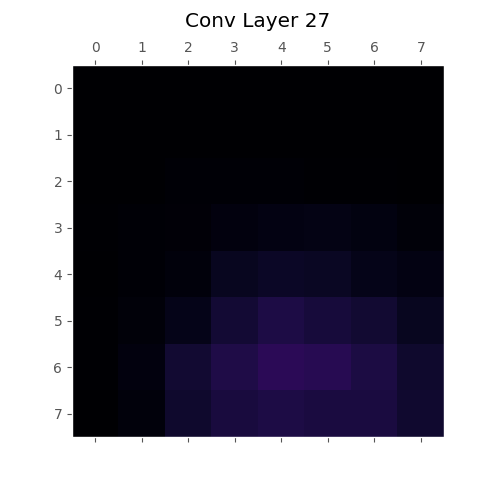}
\includegraphics[width=0.3\columnwidth]{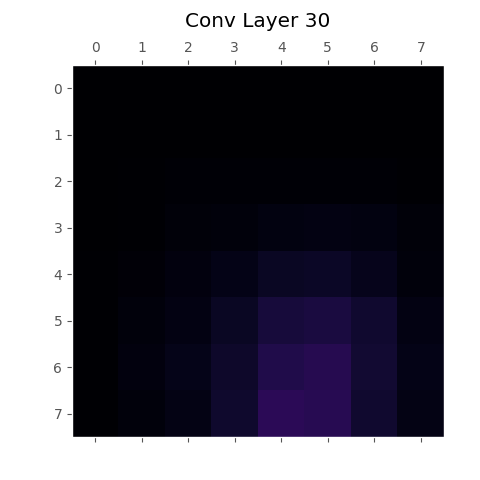}
\includegraphics[width=0.3\columnwidth]{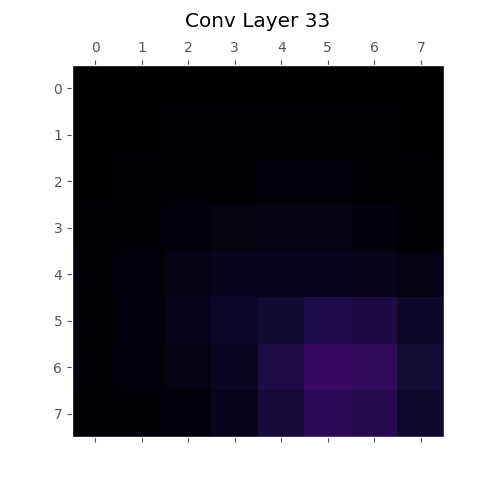}
\caption{Propagation Map (PMavg) for Resnet using a sample from CIFAR.
For this experiment, Equation~\ref{eavg} is used. 
Values are scaled with the maximum value for each layer of the feature maps being the maximum value achievable in the color map.
}
\label{advavg}
\end{center}
\vskip -0.2in
\end{figure*}

Perturbation on the input image propagates throughout the neural network to change its class in adversarial samples.
However, much of this process is unknown.
In other words, how does this perturbation causes a change in the class label? 
What are the internal differences between adversarial attacks and failed attacks if any?

Here, to walk towards an answer to the questions above we propose a technique called propagation maps which can reveal the perturbation throughout the layers.
Propagation maps consists of comparing the feature maps of both adversarial and original samples.
Specifically, by calculating the difference between the feature maps and averaging them (or getting their maximum value) for each convolutional layer, the perturbation's influence can be estimated. 
Consider an element-wise maximum of a three dimensional array $O$ for indices $a$, $b$ and $k$ to be described as:
\begin{equation}
M_{a,b}= \max_k({O_{a,b,0}, O_{a,b,1},...,O_{a,b,k}}),
\end{equation}
where $M$ is the resulting two dimensinal array.

Therefore, for a layer $i$, its respective propagation map $PM_i$ can be obtained by:
\begin{equation}
PM_i = \max_k(FM_{i,k} - FM_{i,k}^{adv}),
\label{emax}
\end{equation}
where $FM_{i,k}$ and $FM_{i,k}^{adv}$ are respectively the feature maps for layer $i$ and kernel $k$ of the natural (original) and adversarial samples.

Alternatively, one may wish to see the average over the filters which exposes a slightly different influence diluted over the kernels in the same layer.
It can be computed as follows:
\begin{equation}
PM_i = \frac{1}{nk}\sum_k(FM_{i,k} - FM_{i,k}^{adv}),
\label{eavg}
\end{equation}
where $nk$ is the number of filters.

PMmax and PMavg will be used to differentiate between Propagation Maps using Equations~\ref{emax} and~\ref{eavg}.
Notice that in order to put the perturbation's influence in the same scale of the original feature map, the maximum scale value will be set to the original feature map when plotting.

\section{Propagation Maps for the One-Pixel Attack}

\begin{figure*}[ht]
\vskip 0.2in
\begin{center}
%centerline{
\includegraphics[width=0.4\columnwidth]{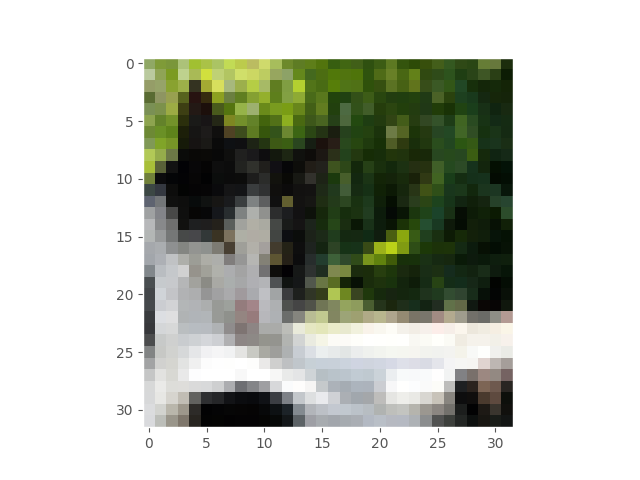}
\includegraphics[width=0.3\columnwidth]{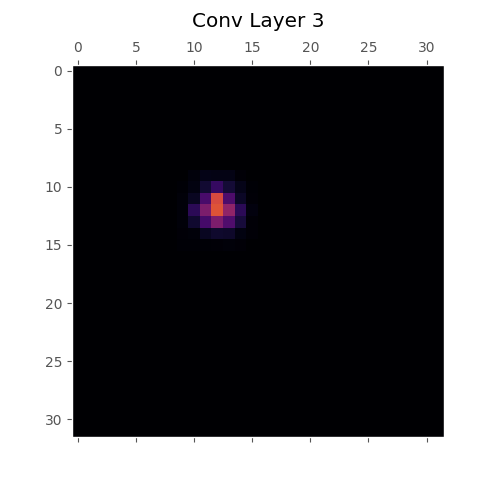}
\includegraphics[width=0.3\columnwidth]{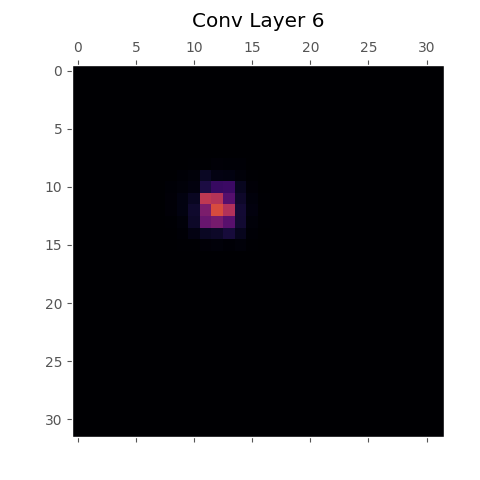}
\includegraphics[width=0.3\columnwidth]{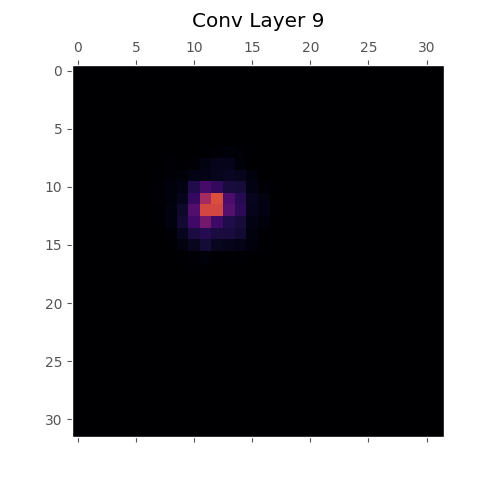}
\includegraphics[width=0.3\columnwidth]{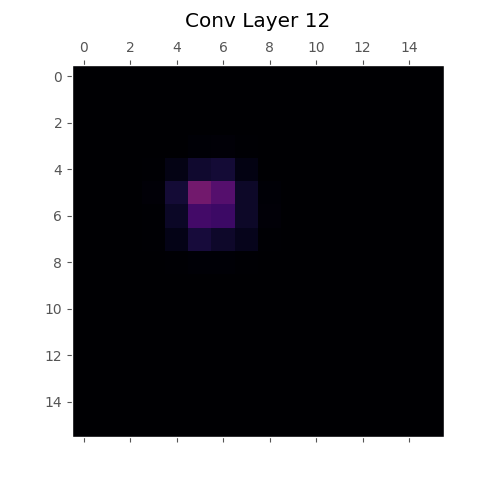}
\includegraphics[width=0.3\columnwidth]{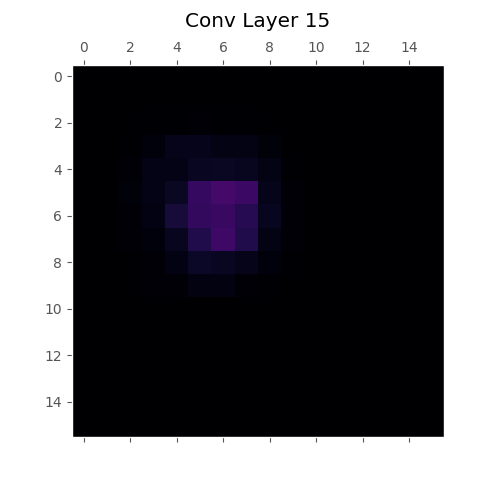}
\includegraphics[width=0.3\columnwidth]{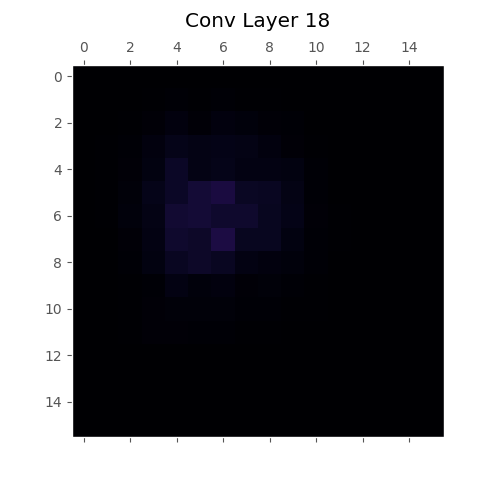}
\includegraphics[width=0.3\columnwidth]{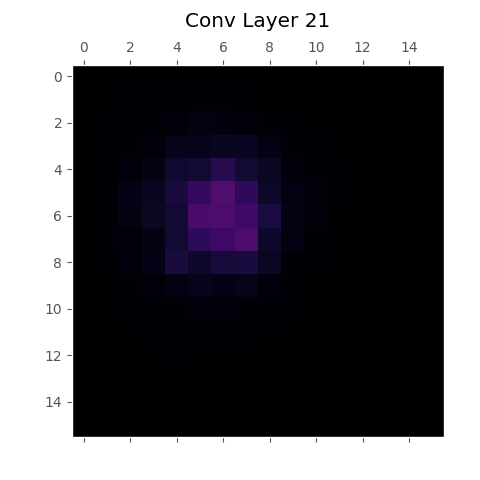}
\includegraphics[width=0.3\columnwidth]{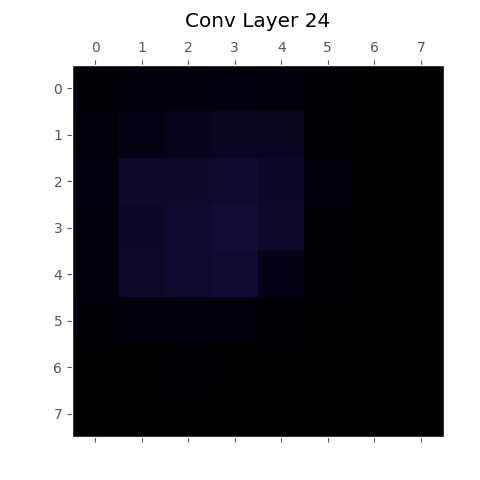}
\includegraphics[width=0.3\columnwidth]{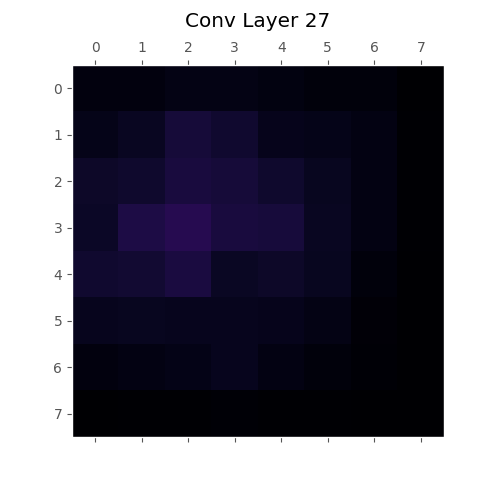}
\includegraphics[width=0.3\columnwidth]{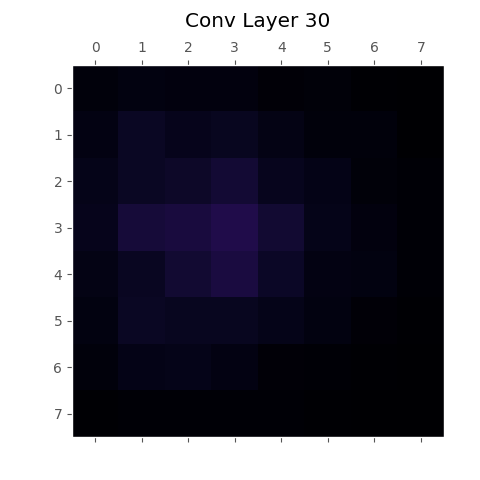}
\includegraphics[width=0.3\columnwidth]{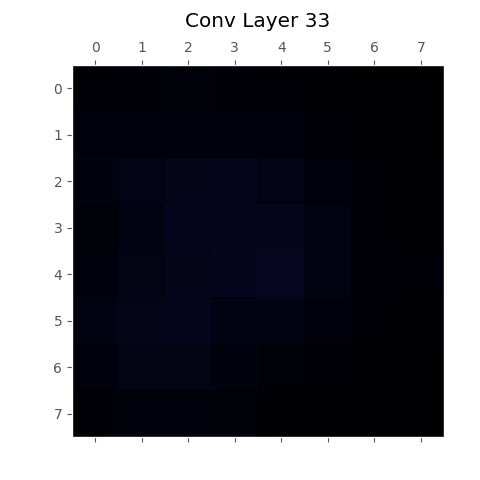}
\caption{Propagation Map (PMmax) for a perturbation that failed to change the class for Resnet using a sample from CIFAR.
For this experiment, Equation~\ref{emax} is used. 
The sample above is \textbf{correctly} classified as cat even after one pixel is changed in the image.
Values are scaled with the maximum value for each layer of the feature maps being the maximum value achievable in the color map.
%Values are scaled with the maximum value for each layer of the feature maps being the maximum value achievable in the color map.
}
\label{missmaxcat}
\end{center}
\vskip -0.2in
\end{figure*}

\begin{figure*}[ht]
\vskip 0.2in
\begin{center}
%centerline{
\includegraphics[width=0.4\columnwidth]{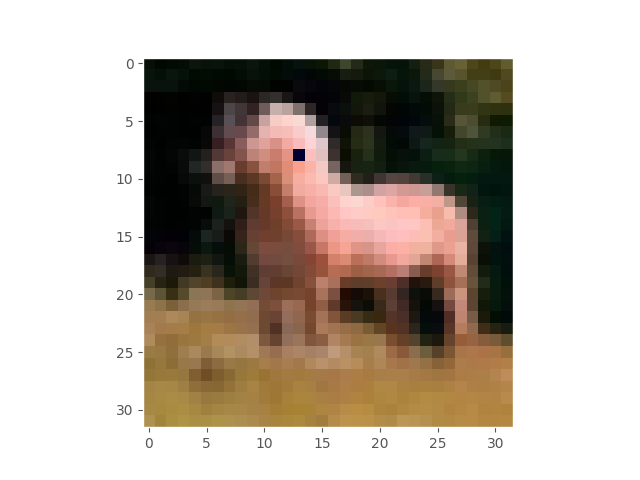}
\includegraphics[width=0.3\columnwidth]{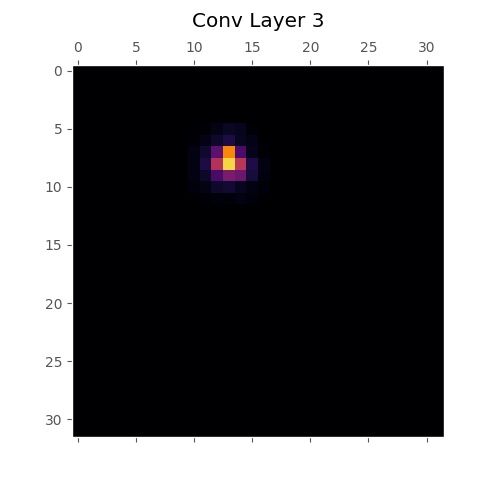}
\includegraphics[width=0.3\columnwidth]{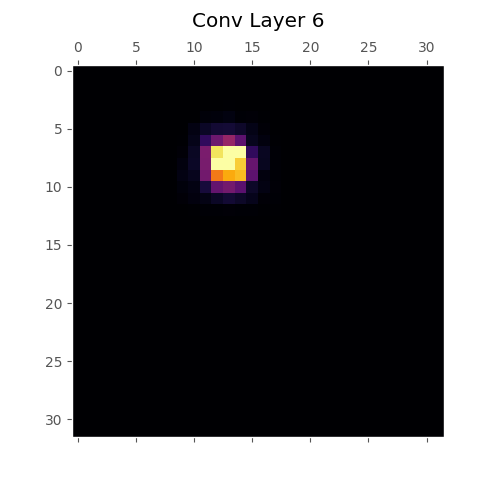}
\includegraphics[width=0.3\columnwidth]{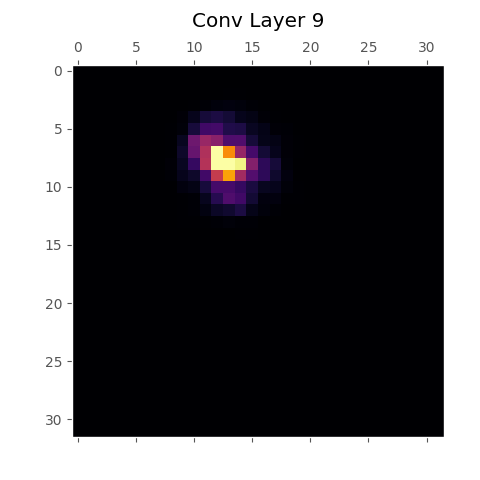}
\includegraphics[width=0.3\columnwidth]{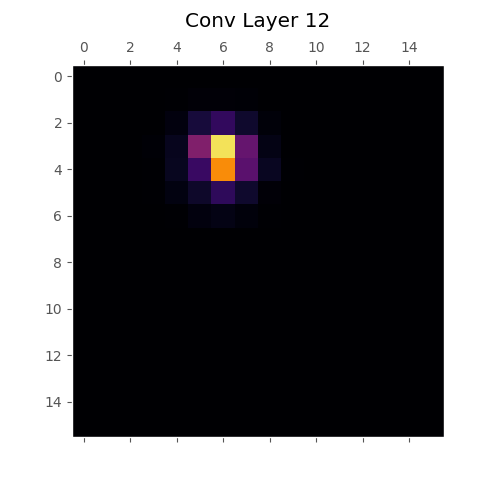}
\includegraphics[width=0.3\columnwidth]{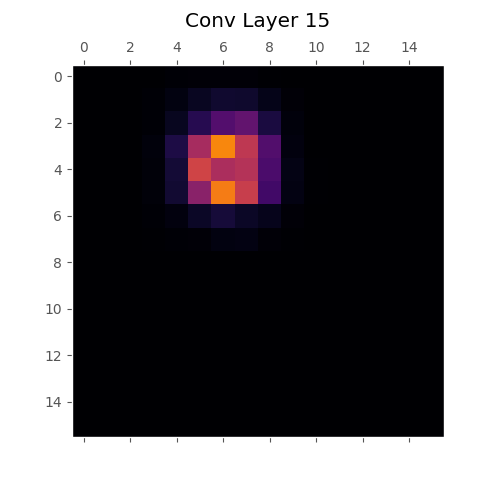}
\includegraphics[width=0.3\columnwidth]{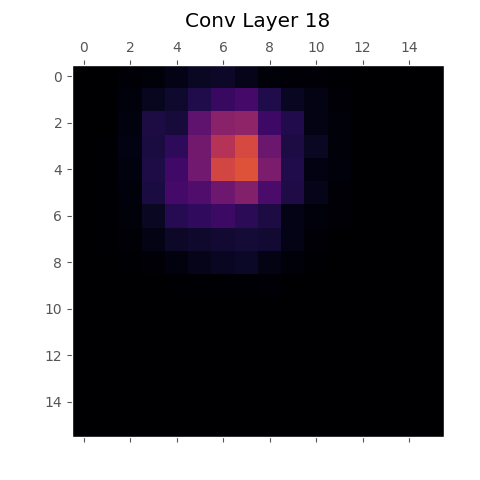}
\includegraphics[width=0.3\columnwidth]{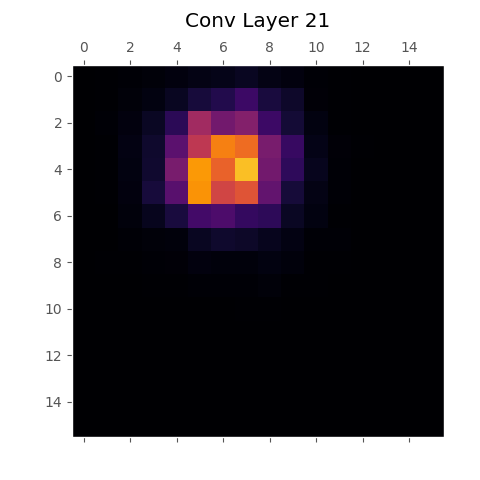}
\includegraphics[width=0.3\columnwidth]{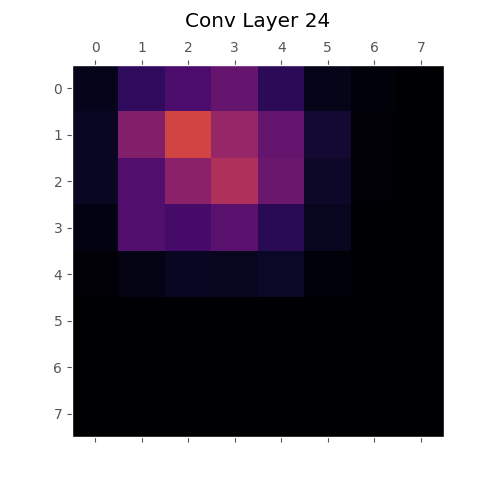}
\includegraphics[width=0.3\columnwidth]{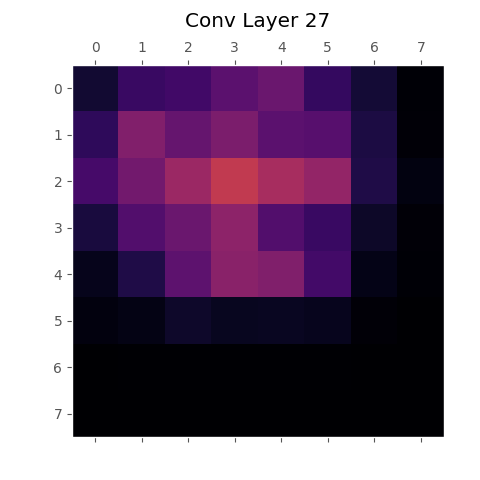}
\includegraphics[width=0.3\columnwidth]{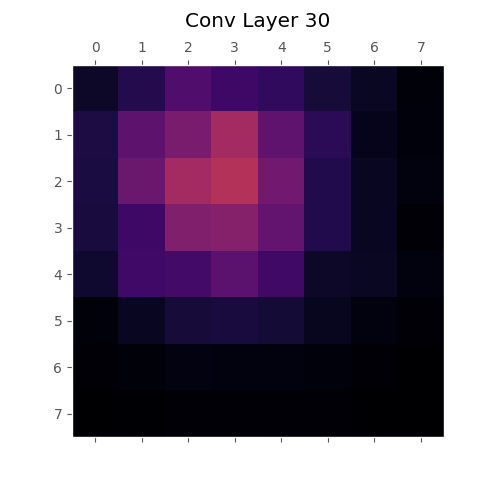}
\includegraphics[width=0.3\columnwidth]{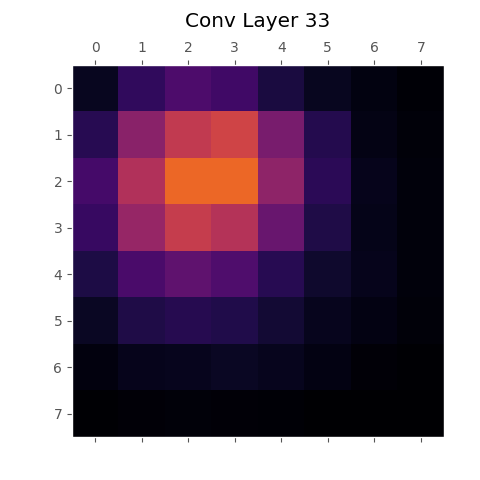}
\caption{Propagation Map (PMmax) for a perturbation that failed to change the class for Resnet using a sample from CIFAR.
For this experiment, Equation~\ref{emax} is used. 
The sample above is \textbf{correctly} classified as horse even after one pixel is changed in the image.
Values are scaled with the maximum value for each layer of the feature maps being the maximum value achievable in the color map.
}
\label{missmaxhorse}
\end{center}
\vskip -0.2in
\end{figure*}

To investigate how a single pixel perturbation can cause changes in class, we will make use of the proposed propagation maps.
This will allow us to vizualize the perturbations in each of the layers of the neural network.

For the experiments, Resnet, which is one of the most accurate types of neural networks, is used.
Each of the subsections below investigate a specific scenario.

\subsection{Single Pixel Perturbations that Change Class}

%Since values are scaled taking as maximum the layer's maximum value, bright colors are close to the maximum value of the entire layer.
%Therefore, the bright colors in the propagation map reaveal a strong influence of the perturbation.

In Figure~\ref{advmax}, the propagation map (PMmax) of a successful one-pixel attack is shown.
The perturbation is shown to start small and localized and then spread in deeper layers.
In the last layer, the perturbation spread enough to influence strongly more than a quarter of the feature map.
This is the element-wise maximum behavior which allows us to identify how strong is the maximum difference in feature maps.

The propagation map based on averaged differences (PMavg) show that the difference is concentraded in some feature maps (Figure~\ref{advavg}).
Moreover this average difference is kept more or less the same throughout the layers.
In the case of PMmax, the difference had a sort of wave behavior, sometimes growing in strength, sometimes slightly fading away.
All observed adversarial samples shared similar features of propagation maps.
This is to be expected, since they need to influence enough in order to change the class.

Surprisingly, one pixel change can cause influences that spread over the entire feature map, specially in deeper layers.
This also contradicts to some extent the expectation that high level features will be processed in deeper layers.

\subsection{Single Pixel Perturbations that Do Not Change Class}

\begin{figure*}[ht]
\vskip 0.2in
\begin{center}
%centerline{
\includegraphics[width=0.4\columnwidth]{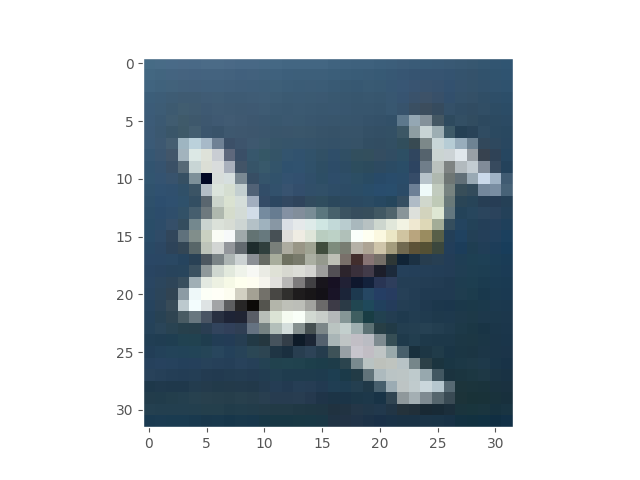}
\includegraphics[width=0.3\columnwidth]{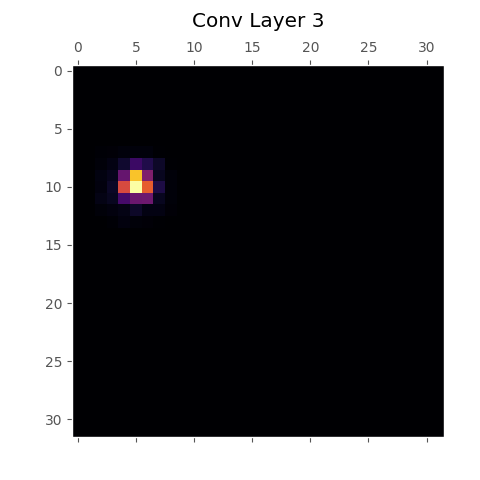}
\includegraphics[width=0.3\columnwidth]{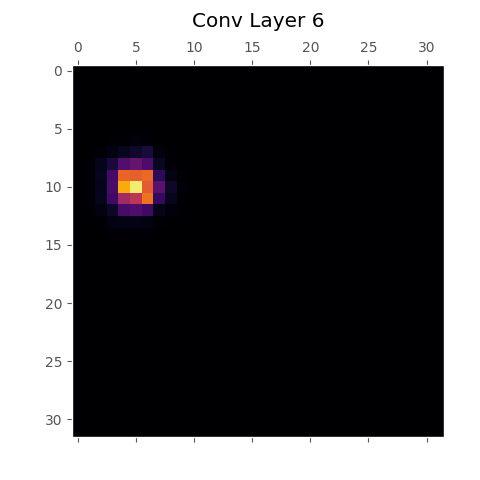}
\includegraphics[width=0.3\columnwidth]{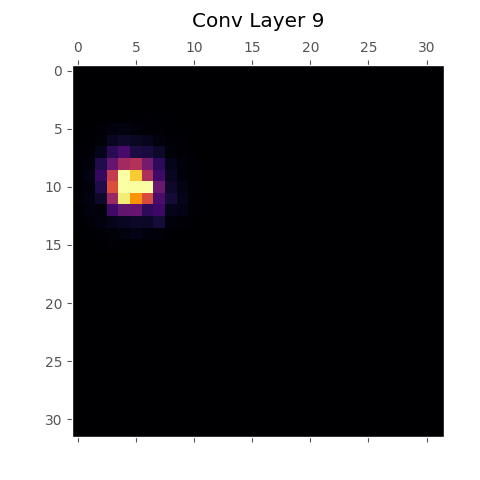}
\includegraphics[width=0.3\columnwidth]{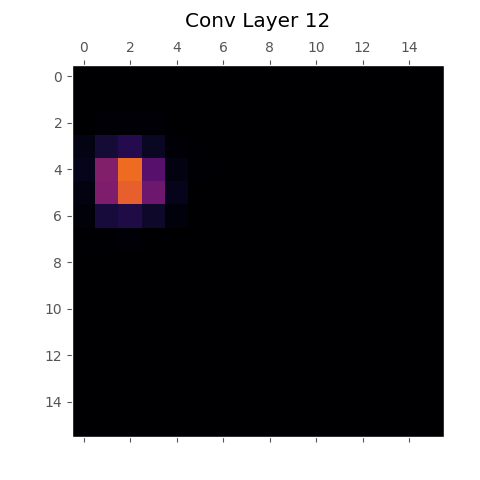}
\includegraphics[width=0.3\columnwidth]{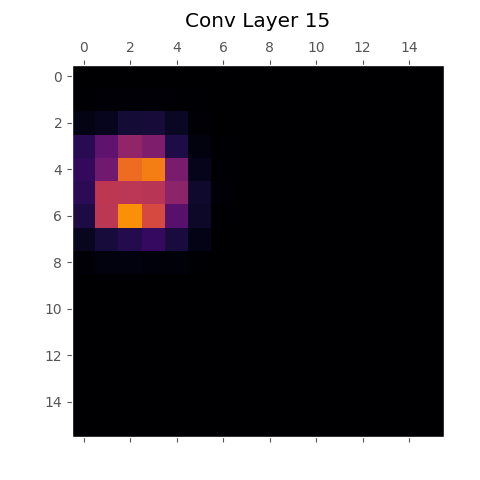}
\includegraphics[width=0.3\columnwidth]{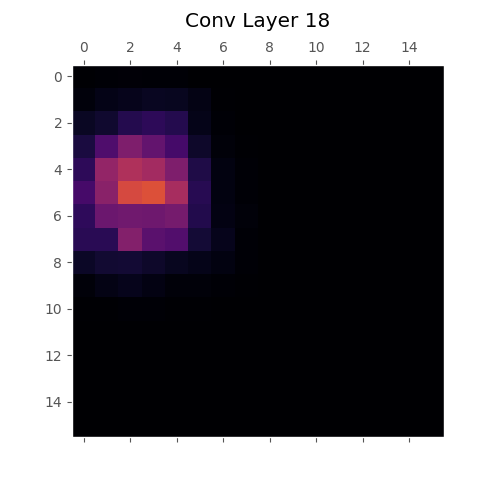}
\includegraphics[width=0.3\columnwidth]{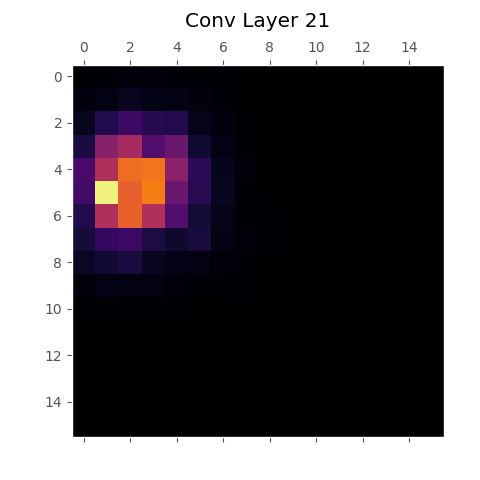}
\includegraphics[width=0.3\columnwidth]{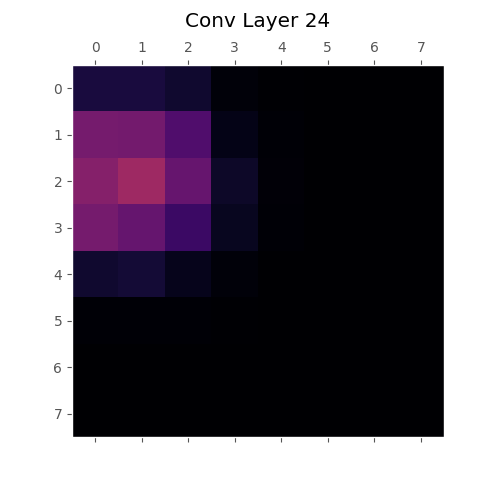}
\includegraphics[width=0.3\columnwidth]{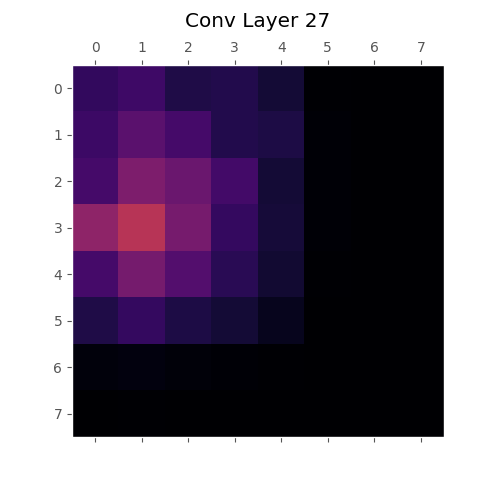}
\includegraphics[width=0.3\columnwidth]{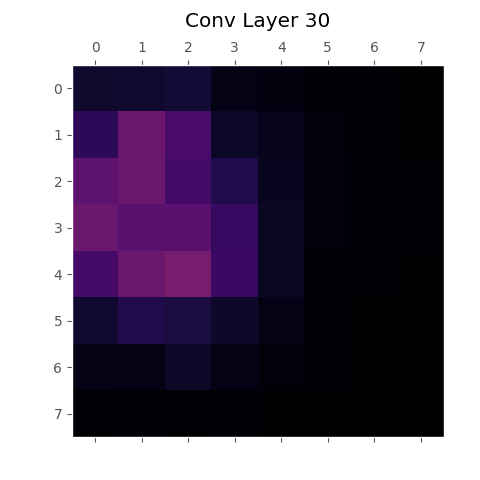}
\includegraphics[width=0.3\columnwidth]{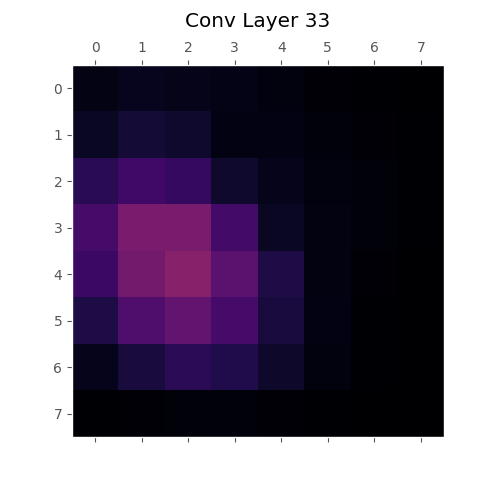}
\caption{Propagation Map (PMmax) for a perturbation that failed to change the class for Resnet using a sample from CIFAR.
For this experiment, Equation~\ref{emax} is used. 
The sample above is \textbf{correctly} classified as airplane even after one pixel is changed in the image.
Values are scaled with the maximum value for each layer of the feature maps being the maximum value achievable in the color map.
}
\label{missmaxair}
\end{center}
\vskip -0.2in
\end{figure*}

Successful one-pixel attacks were shown to grow its influence throughout the layers, culminating in a strong and spread influence in the last layers.
Here we change the position of the pixel to unable the attack to succeed.
Figure~\ref{missmaxcat} shows that in such a case the influence's intensity decreases.
In fact, in the last layer it is almost imperceptible the influence.
However, this is not the rule.
A counterexample is shown in Figure~\ref{missmaxhorse} in which a pixel is changed without changing the class label.
This time however, the perturbation propagates strongly, being as strong if not stronger than the successful one-pixel attack observed in Figure~\ref{advmax}.
One might argue that the influence has caused the confidence to decrease but not enough to cause the change.
For this case, indeed the confidence decreases from $99\%$ to $52\%$.
However, Figure~\ref{missmaxair} has a similar behavior although the confidence decreases only one percentage (from $100\%$ to $99\%$). 
%It can be argued that Figure~\ref{missmaxhorse} has relatively a much stronger influence than this latter one but this are all qualitative statements.

Thus, qualitatively speaking, unsuccessful one-pixel attacks not necessarily fail to achieve a high influence in the last layer.
It depends strongly on the pixel position and sample.
This is accordance with saliency maps which show that different parts of the image have different importance in the recognition process \cite{simonyan2013deep}.

\section{Statistical Evaluation of Propagation Maps}

In previous sections, single attacks were analyzed in the light of examples and counterexamples.
%, showing what happens in these single attacks.
These experiments were important to investigate what happens in detail for each of these attacks.
However, they do not tell much about the distribution of attacks.
This section aims to fill this gap by investigating the attacks' distribution and other statistically relevant data.

\begin{figure*}[ht]
\vskip 0.2in
\begin{center}
%centerline{
\includegraphics[width=0.3\columnwidth]{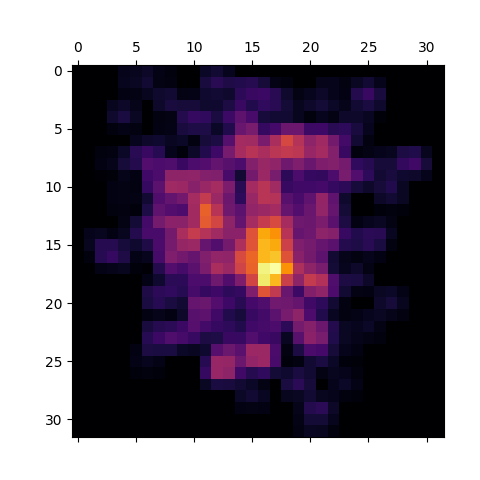}
\includegraphics[width=0.3\columnwidth]{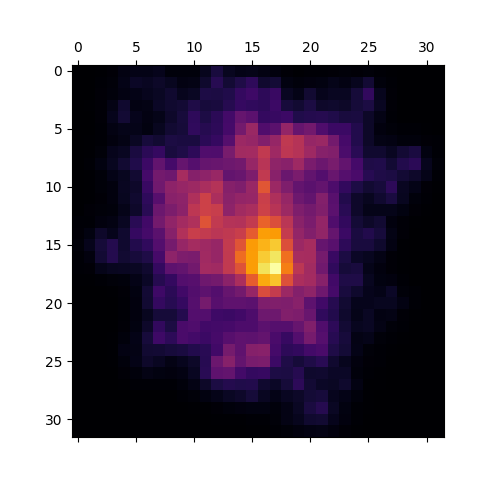}
\includegraphics[width=0.3\columnwidth]{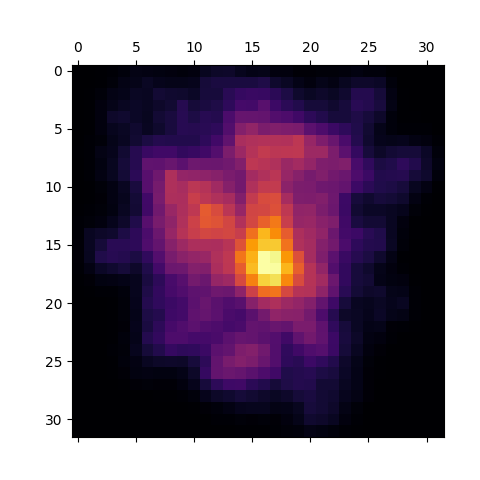}
\includegraphics[width=0.3\columnwidth]{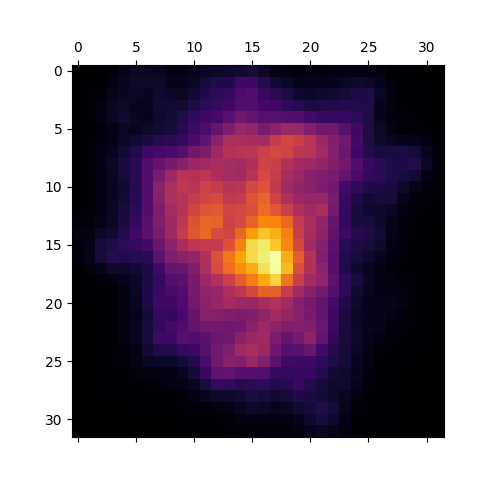}
\includegraphics[width=0.3\columnwidth]{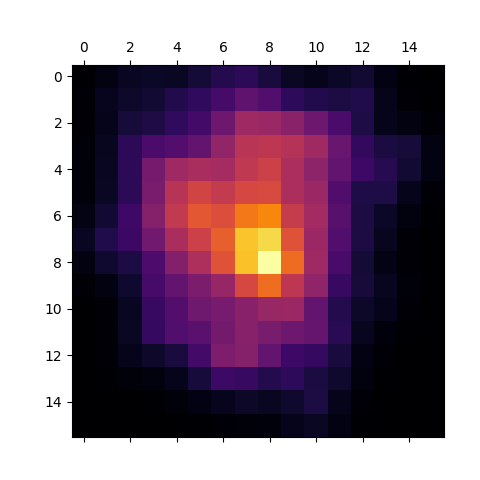}
\includegraphics[width=0.3\columnwidth]{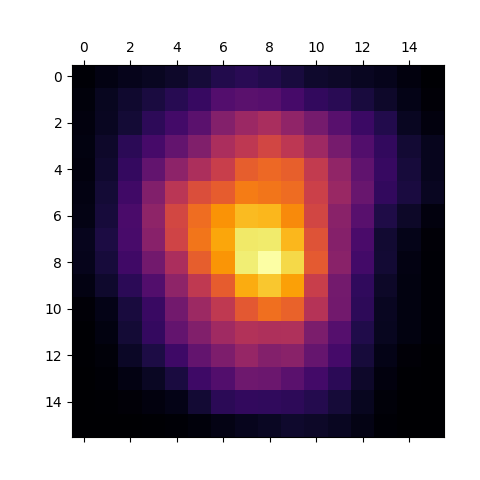}
\includegraphics[width=0.3\columnwidth]{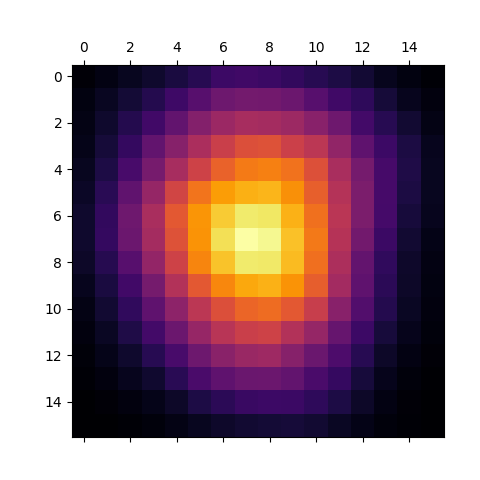}
\includegraphics[width=0.3\columnwidth]{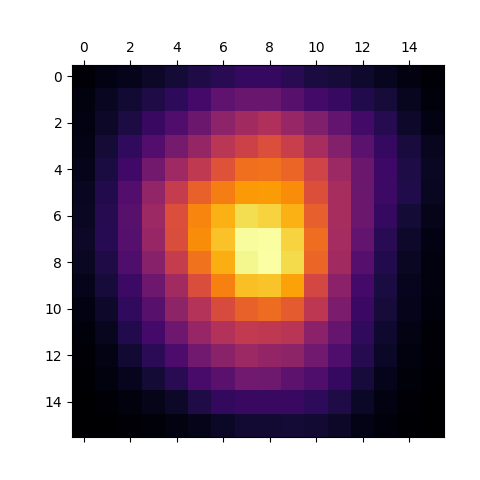}
\includegraphics[width=0.3\columnwidth]{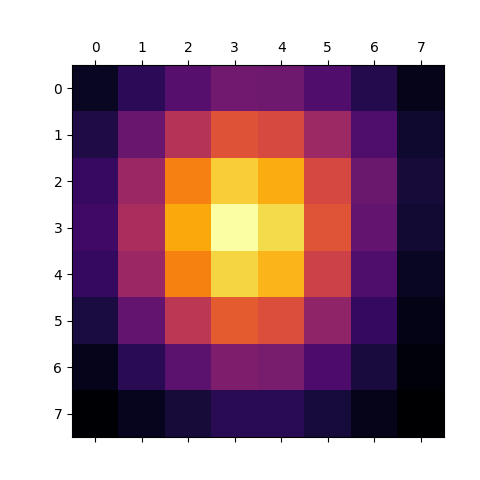}
\includegraphics[width=0.3\columnwidth]{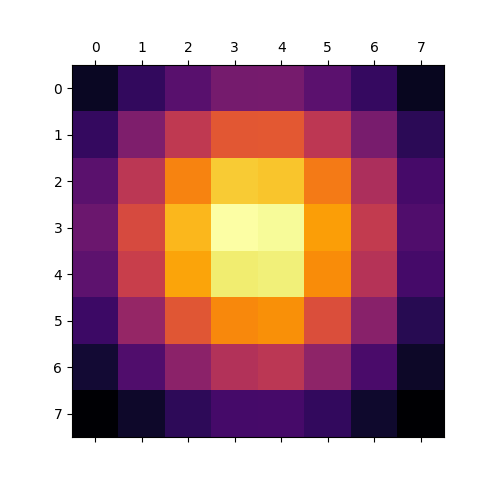}
\includegraphics[width=0.3\columnwidth]{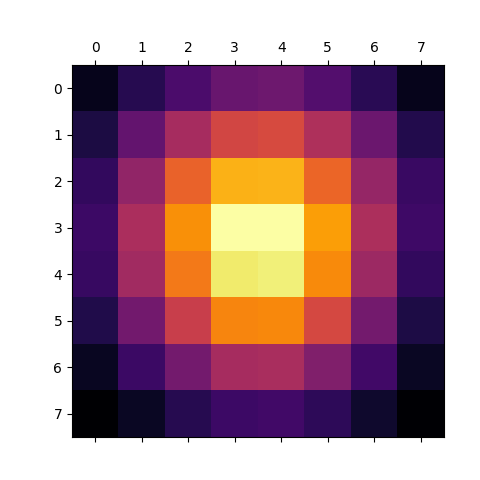}
\includegraphics[width=0.3\columnwidth]{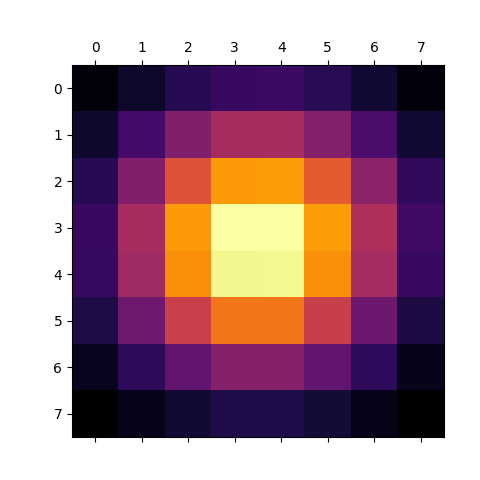}
\caption{
Average of PMmean over $318$ successful attacks on Resnet from CIFAR dataset, i.e., all successful attacks from $1000$ trials.
}
\label{meanpm_hit}
\end{center}
\vskip -0.2in
\end{figure*}

\begin{figure*}[ht]
\vskip 0.2in
\begin{center}
%centerline{
\includegraphics[width=0.3\columnwidth]{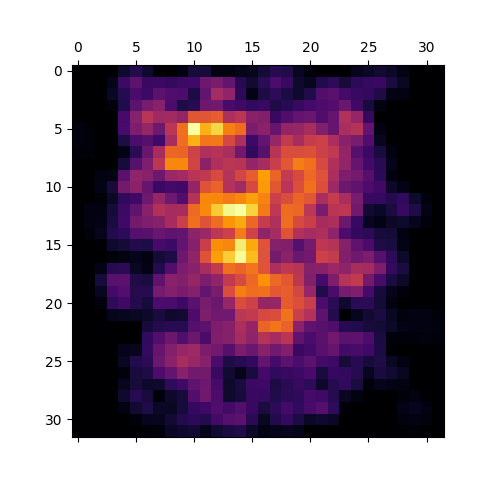}
\includegraphics[width=0.3\columnwidth]{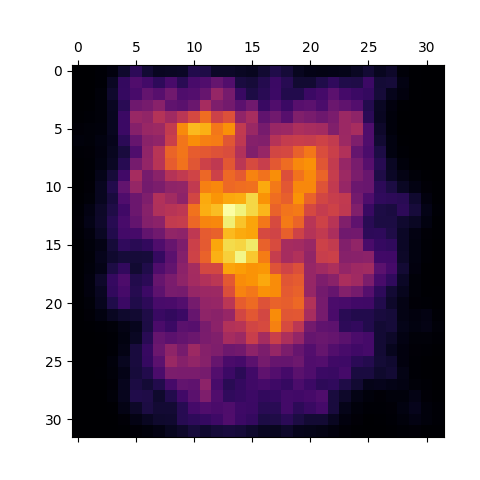}
\includegraphics[width=0.3\columnwidth]{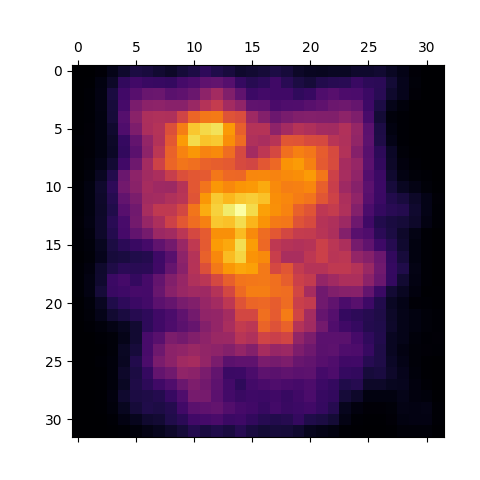}
\includegraphics[width=0.3\columnwidth]{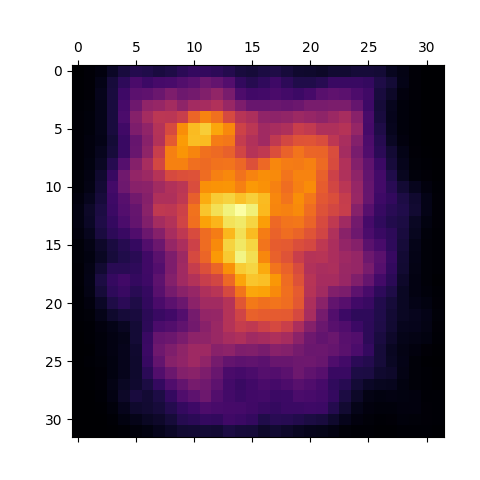}
\includegraphics[width=0.3\columnwidth]{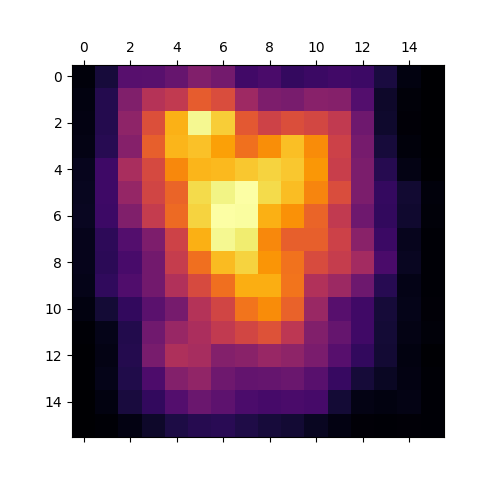}
\includegraphics[width=0.3\columnwidth]{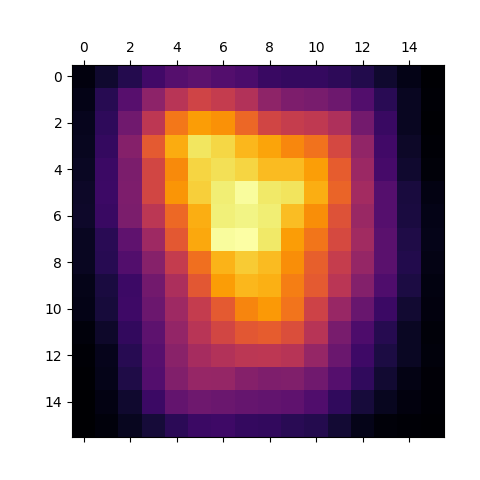}
\includegraphics[width=0.3\columnwidth]{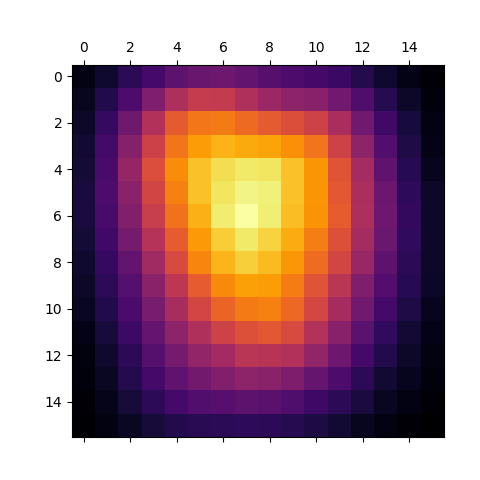}
\includegraphics[width=0.3\columnwidth]{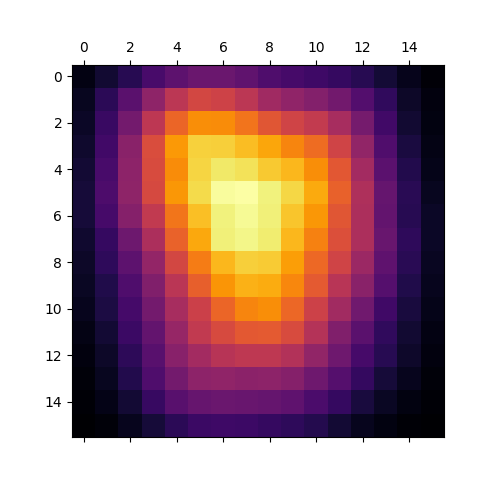}
\includegraphics[width=0.3\columnwidth]{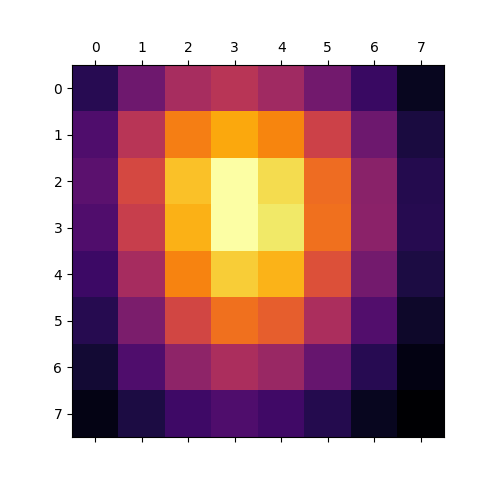}
\includegraphics[width=0.3\columnwidth]{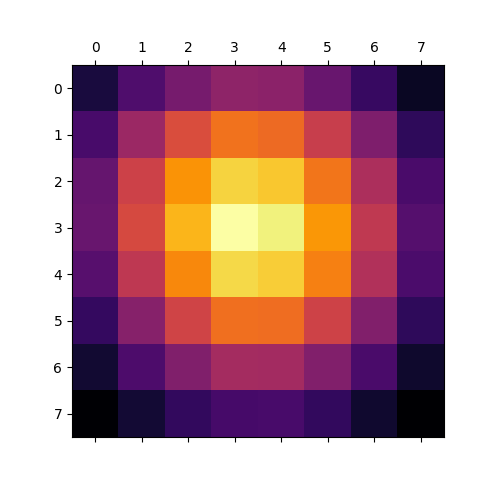}
\includegraphics[width=0.3\columnwidth]{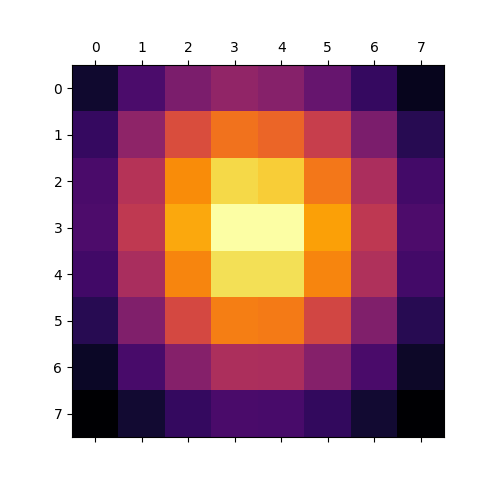}
\includegraphics[width=0.3\columnwidth]{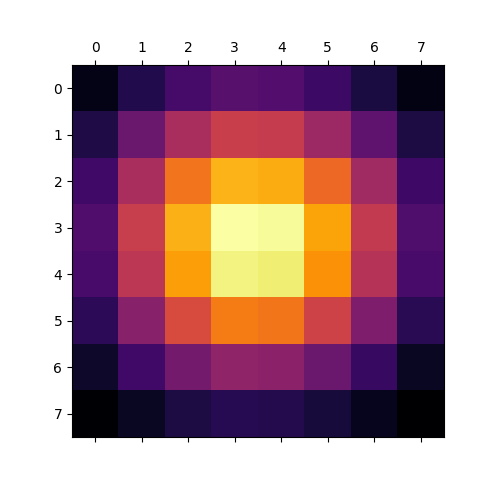}
\caption{
Average of PMmean over $682$ unsuccessful attacks on Resnet from CIFAR dataset, i.e., all failed attacks from $1000$ trials.
}
\label{meanpm_miss}
\end{center}
\vskip -0.2in
\end{figure*}

By evaluating the average of PMmean over successful attacks, we can observe the attack distribution over the entire feature map as well as their average spreadness over the layers  (Figure~\ref{meanpm_hit}).
The successful attacks seem to concetrate mostly close to the center of the image.
In deeper layers, the influence expands and increase in intensity, specially at its center.
This reveals that the behavior observed in Figure~\ref{advmax} is usual for most of the attacks.

Given this distribution for successful attacks, it would be interesting to contrast them with failed attempts.
This would enable us to further clarify the characteristics of a successful attack.
However, as shown in Figure~\ref{meanpm_miss}, most of the previous features are shared.
The only observed difference is the initial spreadness and spreadness.
In other words, failed attacks tend to be farther away from the center of the image and spread more equally than their successful counterparts.
Notice that both Figure~\ref{meanpm_hit} and~\ref{meanpm_miss} are not scaled with the maximum feature map, since we are interested here in their distribution rather than intensity.

\begin{figure}[ht]
\vskip 0.2in
\begin{center}
%centerline{
\includegraphics[width=1.0\columnwidth]{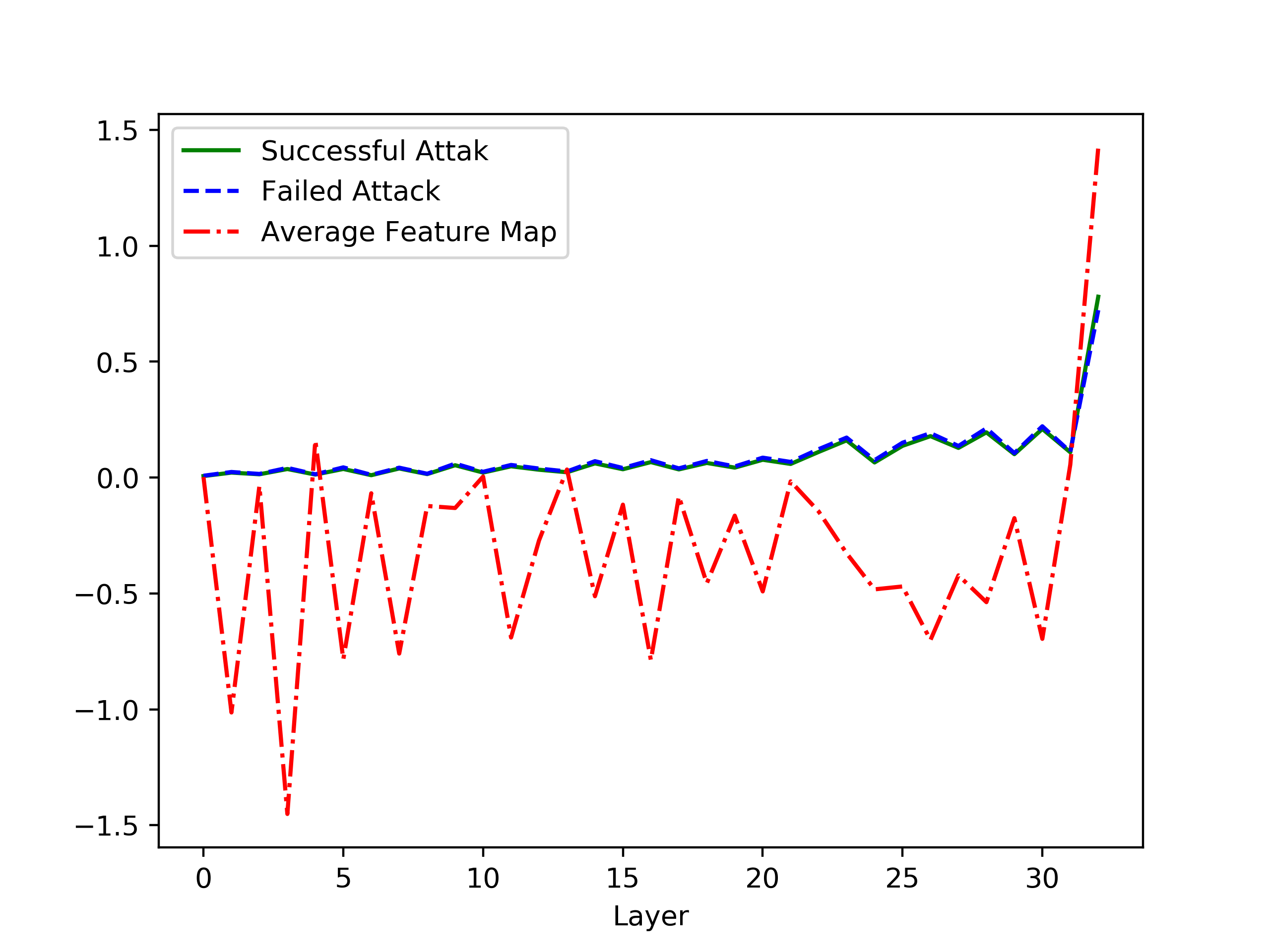}
\caption{
Average difference for all layers when the attack is successful, when it fails and the average value of the feature map without any modification.
}
\label{avglayers}
\end{center}
\vskip -0.2in
\end{figure}

To further clarify if there is any explicit difference between successful and failed attacks, we explictly calculated the mean over all the feature maps for the previous successful and unsuccessful attacks.
The plot is shown in Figure~\ref{avglayers}.
The average difference is also unable to distinghish between successful and failed attacks, with both having very similar behavior.

%\begin{figure*}[ht]
%\vskip 0.2in
%\begin{center}
%centerline{
%\includegraphics[width=1\columnwidth]{statistics_boxplot.png}
%\caption{
%Propagation Map (PMmax) for a perturbation that failed to change the class for Resnet using a sample from CIFAR.
%}
%\label{missmaxair}
%\end{center}
%\vskip -0.2in
%\end{figure*}

\section{Position Sensitivity and Locality}

\begin{table*}[th]
\caption{Success rate of one-pixel attack on both nearby pixels and a single randomly chosen pixel. 
This experiment is conducted in the following manner. 
First the one-pixel attack is executed.
Afterwards, the same perturbation is used to modify one random pixel of the original image and evaluate success of the method. 
To obtain a statistically relevant result, both random and nearby pixel attack are repeated once per image for each successful attack in $5000$ samples of the CIFAR-10 dataset (in which there are $1638$ successful attacks for Resnet and $2934$ successful attacks for Lenet).
}
\label{random}
\vskip 0.15in
\begin{center}
\begin{small}
\begin{sc}
\begin{tabular}{lcccr}
\toprule
 & Lenet & Resnet \\
\midrule
Original One-pixel Attack    & 59\% & 33\%  \\
%One-Pixel Attack on {\bf Random Pixels} & 2.0\% & 0.8\% \\
One-Pixel Attack on {\bf Random Pixels} & 4.9\% & 3.1\% \\
One-pixel Attack on {\bf Nearby Pixels} & 33.1\% & 31.3\%  \\
\bottomrule
\end{tabular}
\end{sc}
\end{small}
\end{center}
\vskip -0.1in
\end{table*}

The one-pixel attack works by searching for a one-pixel perturbation where the class can be modified (misclassified).
This search process is costly but to what extent is the success of the attack dependent on its position?

Here the position sentitivity of the attack will be analyzed.
First, we consider an attack in which a pixel is randomly chosen to be perturbated by the same ammount that could create an adversarial sample.
The results, which are shown in Table~\ref{random}, demonstrate that random pixel attacks have a very low success rate.
This suggests that position is important and by disregarding it is almost impossible to achieve a successful attack.
%Having said that, a random one-pixel attack \footnote{It is important to note that the random pixel is chosen once and therefore this is not a random search procedure.} would also be possible and is consistent for all architectures tested.
Having said that, the attack on nearby pixels (i.e., the eight adjacent pixels) shows a positive result.
%ery high accuracy for a random attack, i.e. there is no search.
%This test shows a decrease of less than $50\%$ from the original one-pixel attack (on average the success rate of the attacks is $54\%$ of the one-pixel one). 
%Contrary to the previous test in which success rate decreases by an order of magnitute, this is a very positive result.
In fact, if we consider that this attack is not conducting any search at all but only taking one random pixel. 
The results may be considered extremely positive.

The extremely positive results present in Table\ref{random}, however, are in accordance with the receptive fields of convolutional layers.
In other words, every neuron in convolutional layers calculates a convolution of the kernel with a part of the input image which is of the same size.
The convolution itself is a linear function in which the change in one input would cause the whole convolution to be affected.
Thus, the result of the convolution will be the same for nearby neurons in the same receptive field. 
Consequently, this shows that the vulnerable part of DNNs were neither neurons nor pixels but some receptive fields.
%Figure~\ref{receptive} illustrate the reasoning.

Interestingly, completely different networks have a very similar success rate for both nearby attacks. % and random attacks on successful attacks.
This further demonstrate that the receptive field is the vulnerable part.
Since neural networks with similar architectures share similar receptive field relationships, nearby attacks on similar networks should have similar success rate.

\section{The Conflicting Salience Hypothesis}

Propagation maps demonstrated that some pixels' influence failed to reach the last layer (Figure~\ref{missmaxcat}) while others influenced the last layer enough to cause a change in class (Figure~\ref{advmax}).
This analysis share a close resemblance to saliency maps in which one wishes to discover which pixels are responsible for a class.
In fact, since propagation maps measure the ammount of influence from perturbations, it would be reasonable to assume that they may have a close relationship with disturbance in saliency maps.
Consequently, adversarial samples would cause enough disturbance in saliency maps to cause a change in class.
Thus, we raise here the hypothesis of a conflicting saliency from adversarial samples.

If this is true, then what adversarial machine learning is doing is not fooling DNNs but rather taking away his attention.
It is like a magician that calls attention to his right hand while his left hand pushes the magic ball.
Or like the blinking light on the street that calls attention of the driver which suddenly drive through a red traffic light.

Having said that, propagation maps is a feedfoward based technique to vizualize and measure the influence of a perturbation while saliency maps aim to investigate the salient pixels for a given class with backpropagated gradients.
Therefore, the methods differ in many ways and their relationship, which might be more complex than what is stated here, goes beyond the scope of this paper.
We leave this as an open question that should be worthwhile to investigate.
%Further investigating their relationship with adversarial samples might shed light into other clues 

\section{Conclusions}
\label{conclusions}

This paper proposed a novel technique called propagation maps and used it to analyze one of the most puzzling attacks, the one-pixel attack.
The analysis showed how a pixel modification causes an influence throughout the layers, culminating in the change of the class.
Moreover, a locality analysis revealed that receptive fields are the vulnerable parts of DNNs and therefore nearby attacks to successful one-pixel attack have a high success rate.
Lastly, a new hypothesis was proposed that could together with the proposed propagation maps help explain the reason behind adversarial attacks in DNNs.

Regarding this paper achievements, we highlight the following:
\begin{itemize}
\item \textbf{ Propagation Maps} - A novel technique to estimate the influence a perturbation can exert over a layer was proposed. 
Propagation maps make it easy to understand both the ammount as well as the spreadness of influence given a perturbation in each layer of a DNN. 
\item \textbf{Vulnerability of Receptive Fields} - A locality analysis revealed that success rates on nearby pixels of successful one-pixel attacks are equally vulnerable. 
This demonstrate that pixels or neurons are not the vulnerable parts but receptive fields.
\item \textbf{The Influence of One-pixel Perturbation} - Propagation maps reveal for the first time how one-pixel perturbation can influence layers of DNNs.
Tests conducted on Resnet show that the perturbation inside neural networks grow and spread. 
In fact, the difference in feature maps was shown to reach values comparable with the maximum output of the original feature map.
\item \textbf{Similarity between Successful and Failed Attacks} - In this paper, many tests over successful and failed attacks were made. 
However, both types of perturbation shows surprisingly similar behavior.
This was shown to not be necessarily related with a decrease in the confidence with the class label either, i.e., failed attacks that do not change the confidence may also have a high influence in all layers of the DNN and behave similarly to successful attacks.
\item \textbf{The Conflicting Salience Hypothesis} - Inspired by how propagation maps show the influence of pixels throughout the layers, we raised the hypothesis that adversarial samples are disturbances in saliency maps.
Saliency maps share a close resemblance with propagation maps. 
Some propagation maps' results also point to conclusions that are in accordance with saliency maps.
Both methods, however, come from quasi contrary perspectives. 
For example, saliency maps is a backpropagated signal while propagation maps is a feedfowarded one. 
To further verify the hypothesis more tests in their relationship are necessary.
\end{itemize}

Thus, by shedding light into the influence of perturbations inside any type of DNN, we expect that propagation maps shall aid the understanding of other attacks as well as the development of new defenses.
They can also be extended to work together with saliency maps providing yet more clues to uncover the reasoning behind adversarial samples as well as how deep neural networks understand the world.
%Additionally, this could shape the path for an evidence that could confirm or refute the conflicting saliency hypothesis.

\section*{Acknowledgements}

This work was supported by JST, ACT-I Grant Number JP-50166, Japan.

\bibliography{adversarial_machine_learning,../sigproc}
%\bibliography{understanding_one}
\bibliographystyle{icml2019}

\end{document}